\documentclass[11pt]{article}

\usepackage[final]{acl}

\usepackage{times}
\usepackage{latexsym}

\usepackage[T1]{fontenc}

\usepackage[utf8]{inputenc}

\usepackage{microtype}

\usepackage{inconsolata}

\usepackage{graphicx}
\usepackage{array}
\usepackage{tabularx}
\usepackage{graphicx}
\usepackage{longtable}
\usepackage{booktabs}
\usepackage{amsmath}
\usepackage{colortbl}
\usepackage{enumitem}

\newcommand{\grayrow}{\rowcolor[gray]{0.9}}
\newcommand{\sparagraph}[1]{\vspace{3pt}\par\noindent\textbf{#1}}

\usepackage[framemethod=TikZ]{mdframed}
\definecolor{OurColor}{HTML}{36aa70}
\definecolor{UserExampleBg}{HTML}{ffffff}
\definecolor{UserExampleTitle}{HTML}{545f7f}
\newmdenv[
    roundcorner=5pt,
    backgroundcolor=UserExampleBg,
    linecolor=UserExampleTitle,
    outerlinewidth=0.5pt,
    frametitlebackgroundcolor=UserExampleTitle,
    frametitlefont={\bfseries\color{white}},
]{user_example}
\AtBeginEnvironment{user_example}{\setlength{\parindent}{0pt}}

\definecolor{MethodTitle}{HTML}{594a4e} 
\newmdenv[
    roundcorner=5pt,
    backgroundcolor=UserExampleBg,
    linecolor=MethodTitle,
    outerlinewidth=0.5pt,
    frametitlebackgroundcolor=MethodTitle,
    frametitlefont={\bfseries\color{white}},
]{user_method}
\AtBeginEnvironment{user_method}{\setlength{\parindent}{0pt}}

\newcommand{\dataset}{\textsc{PACE}}
\newcommand{\model}{\textsc{PaceMaker}}

\title{PACE: Towards Surfacing Hidden Conflicts in User Requests}

\author{%
 Yoojin Kim$^{1}$\quad Jihyoung Jang$^{2}$\quad Hyounghun Kim$^{1,2}$\\
$^{1}$Department of Computer Science and Engineering, POSTECH\\
$^{2}$Graduate School of Artificial Intelligence, POSTECH\\
\texttt{\{kimyujin1224, jihyoung, h.kim\}@postech.ac.kr}\\
}

\begin{document}
\maketitle

\begin{abstract}

Personalized assistants should not only comply with user requests but also assess whether those requests are appropriate given the user's current circumstances. However, prior work has primarily focused on accurately executing requests, overlooking the need for assistants to account for context and engage in conflict-based refusal. Furthermore, while existing work on conflict or safety detection relies on explicitly provided factors, real-world scenarios often involve implicit factors that must be retrieved from a knowledge base (KB). To this end, we introduce \textbf{P}ersonalized \textbf{A}ssistants for \textbf{C}onflict \textbf{E}valuation (\dataset{}), a dataset for evaluating whether models can identify latent constraints, expressed as egocentric knowledge or events, that render seemingly reasonable user requests inappropriate. \dataset{} pairs user requests grounded in well-defined personas with egocentric KB facts, requiring models to integrate contextual evidence to determine whether a request is conflicting. This implicit retrieval setting hinders the direct association between user requests and conflict-inducing knowledge, making it difficult for existing models to identify relevant user-specific facts. To address this challenge, we further propose \model{}, a multi-agent framework in which specialized agents coordinate across query reformulation, multi-hop graph traversal, and conflict-aware filtering to retrieve contextually decisive evidence. Experiments on \dataset{} evaluate both evidence retrieval quality and conflict decision accuracy, showing that \model{} consistently outperforms existing approaches.\footnote{Our code and dataset are publicly available at \url{https://github.com/p2chp2t/pacemaker}.}

\end{abstract}
\section{Introduction}

Recent advances in large language models (LLMs) have transformed AI assistants from passive information retrieval tools into systems capable of supporting users' real-world decisions and actions~\cite{yao2023react,zhang2024agentic,zhang2025survey,peng-etal-2025-survey,xu2026toward}. To be reliable in such settings, assistants must interpret not only the user's immediate request but also whether the requested action is appropriate given the user's personal circumstances, prior commitments, and surrounding conditions~\cite{kim2025learning,lee2025map,yang2026contextagent}.

\begin{figure*}[t]
  \centering
  \includegraphics[width=.98\textwidth]{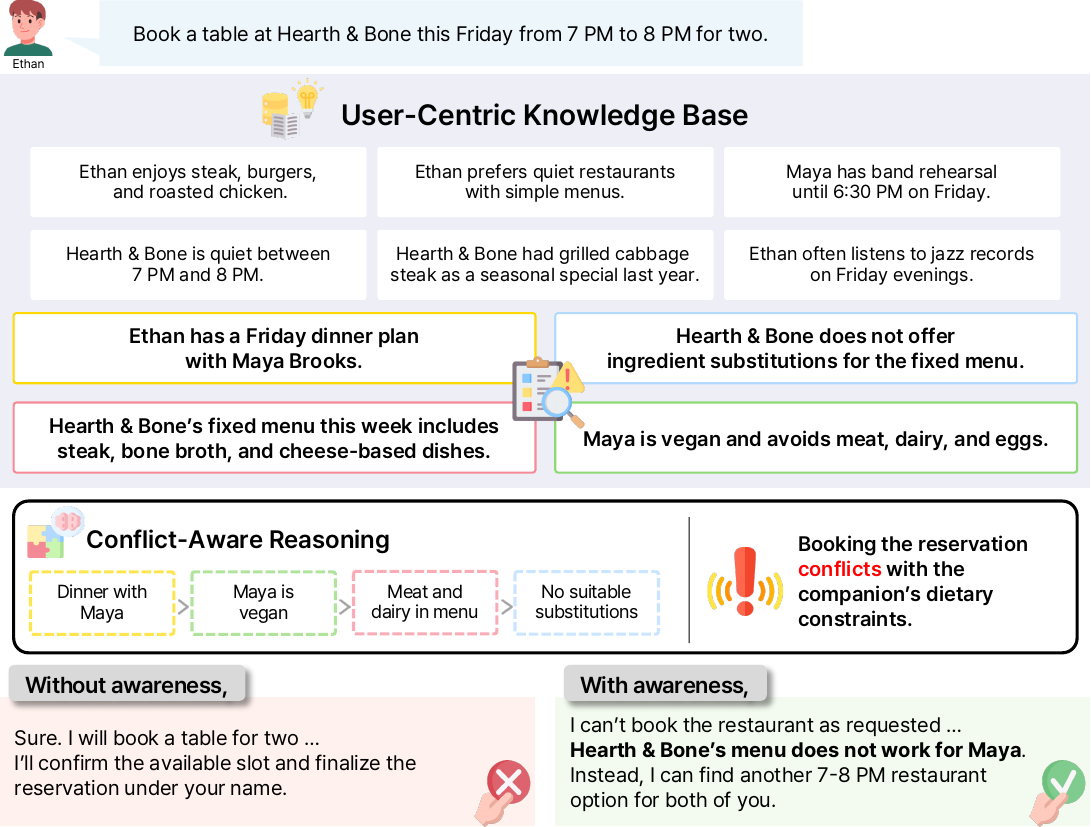}
  \caption{Overview of conflict-aware personalization. The user-centric knowledge base contains information about the user, related people, and surrounding situations. Although the reservation request appears executable, reasoning over evidence about the companion's dietary constraint and the restaurant's fixed menu reveals a latent conflict, requiring the ideal assistant to refuse the booking and suggest an alternative.}
  \label{fig:main}
\end{figure*}

This judgment is often nontrivial. A user request may appear entirely reasonable on its surface yet become inadvisable once hidden constraints are considered~\cite{li2024personal,kim2026propersim}. For example, booking a restaurant seems trivial, but if the user already has a conflicting commitment at that time, or if the venue serves food a companion cannot eat, executing the request without consideration would be inappropriate (Figure~\ref{fig:main}). Conversely, an assistant that over-refuses plausible requests forces users into unnecessary follow-up interactions, degrading both usefulness and efficiency~\cite{rottger-etal-2024-xstest,sun2025casebench,xue-etal-2026-deactivating}. An ideal assistant should therefore make situated decisions, neither blindly executing nor conservatively refusing, but grounding its judgment in evidence relevant to the user's actual circumstances~\cite{zheng-etal-2025-easy}.

Existing benchmarks for safety and risk-aware reasoning have focused on detecting risks or harmful content explicitly present in the input~\cite{mazeika2024harmbench,yuan-etal-2024-r,liu2024mm,andriushchenko2025agentharm,xie2025sorrybench}. While useful for evaluating single-query risk recognition, these input-centric settings fail to capture a critical challenge in real personalized assistant environments. In practice, making appropriate decisions often requires identifying and integrating the handful of facts that truly matter from a large personal knowledge base (KB) filled with plausible but irrelevant distractors~\cite{wu2026personalized}.

To address this gap, we introduce \textbf{P}ersonalized \textbf{A}ssistants for \textbf{C}onflict \textbf{E}valuation (\dataset{}), a retrieval-grounded dataset in which a personalized assistant must reason over an egocentric KB to determine whether a user request should be fulfilled. Each request in \dataset{} is carefully designed to appear as a normal assistant task in isolation; its conflict status emerges only when considered in the context of hidden situational facts distributed throughout the KB. This makes the task particularly difficult, as the decisive evidence is rarely retrievable from the original query alone. Conflict-relevant facts are often distributed and may not appear semantically relevant to the query itself.

Through experiments on \dataset{}, we find that standard retrieval methods struggle to identify contextually decisive evidence in the presence of semantic distractors. Motivated by these findings, we propose \textbf{P}ersonalized \textbf{A}gent for \textbf{C}onflict-\textbf{E}vident \textbf{M}ulti-hop \textbf{A}daptive \textbf{K}nowledge \textbf{E}xtraction (\model{}), a multi-agent framework that reformulates queries to target latent conflict signals, explores evidence through graph-based multi-hop traversal, and filters out distractors to surface only decision-relevant facts. \model{} consistently improves retrieval and reasoning performance over baselines, demonstrating the effectiveness of conflict-aware evidence retrieval.

The contributions of this work are as follows:

\begin{enumerate}
[leftmargin=15pt, itemsep=0pt, topsep=3pt]
\item We introduce \dataset{}, a dataset for evaluating conflict-aware reasoning over egocentric KBs.
\item We propose \model{}, a multi-agent framework for structured evidence retrieval targeting hidden conflict evidence.
\item We conduct experiments showing that surfacing hidden situational constraints remains a substantial open challenge for existing methods.
\end{enumerate}

\section{Related Work}

\sparagraph{Personalized and Context-Aware Assistants.}
Personalization has recently emerged as an important direction for AI assistants~\cite{salemi-etal-2024-lamp,liu2025survey}. Recent work on personalized assistants has examined whether models can understand user-specific information~\cite{tan-etal-2025-personabench}, retain long-term memories~\cite{maharana-etal-2024-evaluating,wu2025longmemeval}, and adapt to evolving user preferences~\cite{jiang2025know,jiang2025personamem,zhao2025do}. In parallel, prior work on contextual safety has shown that the appropriateness of a request can depend on its surrounding situational context~\cite{wang-etal-2025-safe,zhou2025multimodal,son-etal-2025-subtle,lou-etal-2025-think}. However, these settings typically center on preference use or externally observable risks, leaving underexplored cases where inappropriateness is neither explicit in the request nor tied to a single salient fact. We therefore focus on conflict-aware personalization where models must compose distributed egocentric evidence to detect latent constraints on otherwise reasonable requests.

\sparagraph{Retrieval and Memory for Personalized LLMs.}
Existing work on personalized LLMs spans memory systems~\cite{chhikara2025mem0,xu2026amem}, personalized alignment~\cite{li2024personalized,zollo2025personalllm,liang2026learning}, and RAG~\cite{zerhoudi2024personarag}. Recent personalized and graph-based RAG methods improve evidence organization and multi-hop reasoning using user signals~\cite{zhang2026personalize,tan-etal-2025-personabench}, structured personal knowledge~\cite{prahlad2025personalizing}, and graph expansion mechanisms~\cite{edge2024local,guo-etal-2025-lightrag,gutierrez2024hipporag,gutierrez2025from,zhu-etal-2025-knowledge}. Despite these advances, most methods prioritize retrieving relevant or supportive information for answering and personalization, rather than evidence needed for context-sensitive request assessment. Our method instead frames retrieval as diagnostic evidence selection, targeting the facts that determine whether an apparently valid request remains compatible with the user's broader context.

\section{\dataset{}}
We introduce \textbf{P}ersonalized \textbf{A}ssistants for \textbf{C}onflict \textbf{E}valuation (\dataset{}), a new dataset designed to evaluate whether LLMs can recognize situational conflicts between user requests and facts stored in a user-centric KB.

\subsection{Task Definition}
Our goal is to evaluate whether a model can determine if a user request is compatible with facts stored in an egocentric KB. We consider a personalized assistant setting in which the assistant maintains contextual knowledge about a user (i.e., ego) and limited information about closely related individuals such as family, colleagues, or friends (i.e., alters), covering only aspects relevant to the user's decision. Given a request and this KB, the model must decide whether the requested action should be carried out and justify its decision.

A key characteristic of this task is that the request itself appears normal and executable in isolation. The challenge instead arises from hidden situational constraints that become apparent only when the relevant contextual facts in the KB are taken into account.

\subsection{Feasibility Status and Situation Types}
Each instance in \dataset{} consists of a user's requests and an egocentric KB, and each request is annotated with a \textit{feasibility status} indicating whether it is conflicting or non-conflicting under the given KB, and a \textit{situation type} specifying the primary source of reasoning required for the decision.

Conflict cases refer to requests that become inappropriate or incompatible after contextual facts are considered, whereas Non-conflict cases remain feasible and appropriate under the same conditions. We include Non-conflict cases to ensure that models do not simply reject requests whenever contextual information is present.

To enable fine-grained analysis of conflict-aware reasoning, we categorize requests into three situation types: \textit{Temporal}, involving time and schedule constraints; \textit{Personal}, involving preferences or interpersonal constraints; and \textit{State}, involving current conditions and available resources.

\begin{itemize}
[leftmargin=15pt, itemsep=0pt, topsep=3pt]
\item \textbf{Temporal:} This type covers cases where a request is incompatible with temporal constraints such as existing schedules, travel time, or daily routines. For example, if a user asks the assistant to register them for a 19:00 certification exam, but the KB indicates that the user's prior workshop ends at 18:10, their identification documents must be picked up from the hotel by 18:30, and the exam center is 45 minutes away, the request creates a temporal conflict because the connected commitments leave no feasible schedule.

\item \textbf{Personal:} Cases of this type arise when fulfilling a request would significantly violate an important personal constraint of the ego or an alter, such as a health condition, personal value, or accessibility need. For example, if a user asks for a seafood boil restaurant to visit with a friend after an exhibition, but the KB indicates that the friend has a shellfish allergy, recommending such a restaurant would constitute a personal conflict.

\item \textbf{State:} This type covers cases where a request becomes inappropriate due to an external condition already known at query time, such as road conditions, posted restrictions, or facility operating issues. For example, if a user asks for a quiet cafe to work in during the afternoon, but the KB indicates that the cafe they usually visit has scheduled a live music event at that time, recommending that cafe would constitute a state conflict.
\end{itemize}

\subsection{Dataset Generation}
To construct \dataset{}, we first synthesize egocentric persona scenarios and then generate requests and contextual KB facts grounded in those scenarios. The construction process is designed to ensure that (1) requests appear natural and executable in isolation, (2) the final decision depends on hidden contextual constraints, and (3) the required evidence is distributed across multiple KB facts.

\sparagraph{Persona Expansion.}
We begin with persona seeds collected from MSC~\cite{xu-etal-2022-beyond} and Synthetic-Person-Chat~\cite{jandaghi-etal-2024-faithful}. Since the original personas are relatively simple, we use \texttt{GPT-5.4-mini}\footnote{\url{https://openai.com/index/introducing-gpt-5-4-mini-and-nano/}} to expand them into richer narrative descriptions containing everyday routines, living environments, behavioral tendencies, and plausible situational context relevant to personalized assistant interactions.

\sparagraph{Profile Synthesis.}
We then randomly pair the expanded narratives to construct ego-alter relationships. For each pair, \texttt{GPT-5.4-mini} first generates a structured ego profile and subsequently produces an alter profile conditioned on the ego. The final profiles include personal attributes such as occupation, health conditions, and values, along with concrete everyday details involving commonly used places and devices, which later serve as the basis for conflict-aware query and KB generation grounded in realistic life situations.

\sparagraph{Query and Context Generation.}
Based on the synthesized profiles, we use \texttt{GPT-5.4-mini} to generate user requests together with contextual KB facts required for conflict evaluation. Each scenario is constructed within a bounded timeline centered around a reference date to ensure temporal consistency. We treat the reference date as the time of the user request: facts before it are already observed, confirmed, or in effect, while facts after it are included only if they are already scheduled, announced, planned, or otherwise knowable at that time. Each case consists of a query, a gold context, and a judgment.

Queries are designed to resemble ordinary user requests, such as making reservations, planning activities, or providing recommendations, without explicitly revealing the underlying conflict signals. We additionally exclude cases where the request itself appears inherently unreasonable or blatantly inconsistent with the user's established conditions.

The gold context provides the situational background necessary to evaluate the query, typically requiring multiple facts to be interpreted jointly rather than exposing a single decisive clue. It does not directly state the request's feasibility status, so the model must infer the appropriate decision by relating these contextual facts to the user's request.

The judgment provides the reference rationale, explaining why the request should be rejected in Conflict cases or why it remains executable in Non-conflict cases. We further use this judgment to assess whether the model's decision is grounded in the correct evidence and reasoning.

\begin{figure*}[t]
  \centering
  \includegraphics[width=0.98\textwidth]{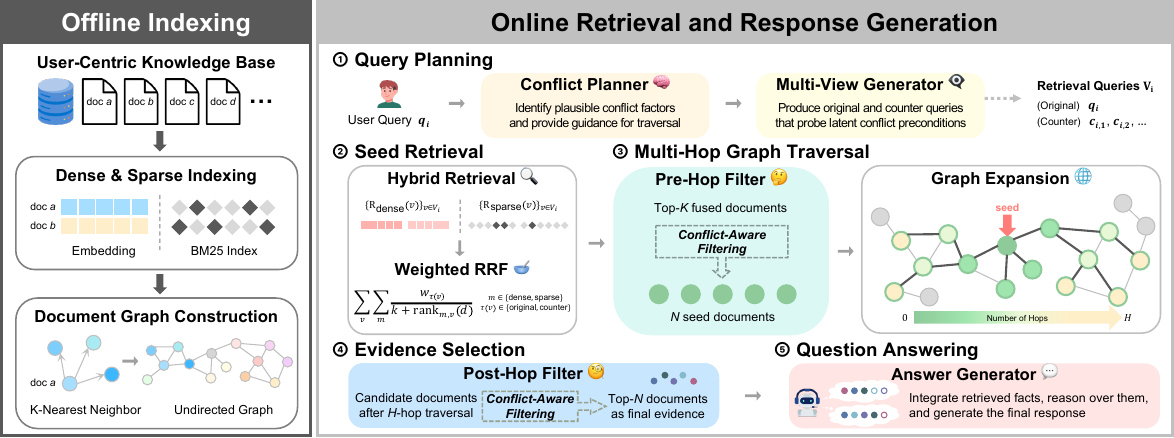}
  \caption{Workflow of \model{}.}
  \label{fig:method}
\end{figure*}

\sparagraph{Distractor Generation and Atomization.}
For each instance, we use \texttt{GPT-5.4-mini} to generate distractor contexts that are topically related to the query but do not reveal the decisive reasoning evidence contained in the gold context. Distractors are constructed to remain consistent with the persona profile, timeline, and intended judgment while avoiding the introduction of additional conflicts or compatibility signals.

Both gold and distractor contexts are further decomposed into atomic facts, which are stored as independent KB entries. This decomposition reflects realistic personal KB structures in which information is stored as discrete and scattered facts rather than coherent passages. The decomposed gold facts do not necessarily contribute equally to the correct judgment. In some cases, a subset suffices, while in others, the full set must be integrated, reflecting the varying complexity of conflict reasoning across instances.

\begin{table}[t]
\centering
\resizebox{0.83\linewidth}{!}{
\begin{tabular}{l c}
\toprule
\textbf{Type} & \textbf{Count}\\
\midrule

\# of facts across all datasets & 376,448\\
\# of queries across all datasets & 3,249\\
\# of profile instances & 185\\

Avg. of facts per instance & 2,035\\
Avg. of queries per instance & 18\\
Avg. of gold facts per query & 4.01\\
\bottomrule
\end{tabular}
}
\caption{Statistics of \dataset{}.}
\label{tab:DataStat}
\end{table}

\sparagraph{Quality Verification.}
Although \dataset{} is synthetic, we ensure its quality through manual review by the authors at each stage of construction, complemented by automated filtering using \texttt{GPT-5.4-mini}. After generating each case, we evaluate it for taxonomy compliance, constraint satisfaction, and core-cause diversity. Cases that satisfy all quality criteria are retained, cases that can be improved are refined based on validator feedback, and cases that fail to meet the quality standards are discarded.

We also verify that the decomposed gold facts are sufficient and unambiguous for resolving each query. Instances that fail this check are excluded, yielding \dataset{} with approximately 3.2K queries (see Table~\ref{tab:DataStat} for statistics). Full process examples and additional statistics are provided in Appendix~\ref{app:dataDetail}, and prompts are provided in Appendix~\ref{app:prompts}.

Furthermore, we perform a human evaluation to validate whether the feasibility status defined by the LLM aligns with human judgments. The results show a high agreement rate of 93.3\%, indicating strong consistency between the generated feasibility labels and human assessments (see Appendix~\ref{app:feasibilityHumanVal} for details).

\section{\model{}}
\label{sec:method}

We propose \textbf{P}ersonalized \textbf{A}gent for \textbf{C}onflict-\textbf{E}vident \textbf{M}ulti-hop \textbf{A}daptive \textbf{K}nowledge \textbf{E}xtraction (\model{}), a multi-agent retrieval framework for conflict-aware reasoning over egocentric KBs. From a user-specific KB consisting of thousands of atomic fact sentences, \model{} first constructs dense and sparse indexes and a k-nearest neighbor (k-NN) document graph. At inference time, given a user query, it clarifies a compact set of conflict-diagnostic documents through four stages: conflict-aware query planning, hybrid retrieval with fusion, agentic multi-hop graph traversal, and conflict-aware document filtering (see Figure~\ref{fig:method}).

\subsection{Conflict-Aware Query Planning}
A single user query may not sufficiently expose the contextual evidence required for conflict detection, particularly when the relevant KB facts do not share lexical overlap with the request. To address this, we employ a two-step query reformulation process that first plans the retrieval direction, and then generates multiple retrieval-oriented views.

Given a user query and a reference date, a \textbf{conflict planner agent} first identifies up to three decision-relevant probing cues, each capturing a potential dimension of conflict such as scheduling constraints, prior commitments, or resource availability. This plan is then passed to a \textbf{multi-view query generator}, which produces (1) the original query view and (2) a set of counter views that explicitly target potentially conflicting conditions implied by the plan. Counter views are designed to retrieve latent conflict evidence that may not be directly implied by the original query. All views are generated conditioned on the reference date to correctly resolve temporal expressions.

\subsection{Hybrid Retrieval and Fusion}
Each query view is passed to both dense and sparse retrievers. The dense retriever encodes queries and KB documents and performs approximate nearest neighbor search, while the sparse retriever applies BM25 over tokenized documents to capture lexical overlap.
Retrieval results from all query views and retrievers are merged using Weighted Reciprocal Rank Fusion (WRRF), where counter-view results are assigned higher weights to prioritize conflict-relevant evidence. The top-$K$ fused documents are then passed through a \textbf{pre-hop filter agent}, which selects the top-$N$ most decision-relevant documents as the seed set for the subsequent traversal stage. This pre-hop filtering step reduces noise before graph traversal begins and prevents misleading seed documents from propagating into larger expansions.

\subsection{Multi-Hop Graph Traversal}
A key challenge in conflict-aware reasoning is that critical evidence is often not directly retrievable from the original query. Relevant evidence may be only weakly related to the query itself but located near the filtered seed documents in the pre-constructed k-NN graph. To surface such evidence, we perform multi-hop traversal over this graph. The filtered seed documents serve as entry points, and traversal proceeds via breadth-first search (BFS), iteratively expanding each frontier document to its top-$M$ neighbors to collect additional contextually related evidence up to a maximum depth of $H$.

\subsection{Conflict-Aware Evidence Selection}
After traversal, the collected document pool, comprising both seed documents and documents discovered through multi-hop traversal, may still contain topically related but weakly informative documents. A \textbf{post-hop filter agent} performs a final evidence selection step over the entire pool, retaining only the top-$N$ documents that most directly contribute to determining whether the request is feasible or conflicts with the user's KB.

\subsection{Answer Generation}
The final filtered evidence set is passed to an \textbf{answer generator}, which produces a final response indicating whether the request can be fulfilled and explaining the reasoning behind the decision. Full implementation details, including hyperparameter configurations and selection, are provided in Appendix~\ref{app:implDetail}. Please refer to Appendix~\ref{app:prompts} for the instructions used by each agent.

\section{Experiments}

\begin{table*}[t]
\centering
\resizebox{0.97\linewidth}{!}{
\begin{tabular}{l c c c c c c c c c c}
\toprule
\textbf{Method}
& \multicolumn{7}{c}{\textbf{Retrieval Performance.}} 
& \multicolumn{3}{c}{\textbf{Response Quality.}} \\
\cmidrule(lr){2-8}\cmidrule(lr){9-11}
& \textbf{Recall@5} & \textbf{Recall@10} & \textbf{Hit@5} & \textbf{Hit@10} & \textbf{Gold@5} & \textbf{Gold@10} & \textbf{MRR}
& \textbf{\textsc{Pass}} & \textbf{\textsc{Wrong}} & \textbf{\textsc{Fail}} \\
\midrule
\grayrow\multicolumn{11}{c}{\textbf{Qwen3-Embedding-8B / Qwen3-4B-Instruct-2507}} \\
\midrule
Oracle
& - & - & - & - & - & - & -
& 86.89 & 1.69 & 11.42 \\
Full KB
& - & - & - & - & - & - & -
& 57.49 & 12.10 & 30.41 \\
Sparse
& 19.77 & 26.52 & 51.45 & 62.61 & 2.35 & 4.90 & 37.01
& 62.73 & 8.13 & 29.15 \\
Dense
& 17.90 & 25.16 & 48.13 & 60.35 & 1.95 & 4.07 & 32.42
& 62.39 & 8.40 & 29.21 \\
\model{}
& 26.23 & 29.51 & 62.41 & 68.98 & 2.83 &  3.01 & 50.14
& 68.82 & 7.60 & 23.58 \\
\midrule
\grayrow\multicolumn{11}{c}{\textbf{text-embedding-3-small / GPT-5.4-mini}} \\
\midrule
Oracle
& - & - & - & - & - & - & -
& 87.13 & 2.09 & 10.77 \\
Full KB
& - & - & - & - & - & - & -
& 73.10 & 7.29 & 19.61 \\
Sparse
& 19.77 & 26.52 & 51.45 & 62.61 & 2.35 & 4.90 & 37.01
& 65.53 & 7.05 & 27.42 \\
Dense
& 14.52 & 21.32 & 39.72 & 52.71 & 1.50 & 3.18 & 27.55
& 62.63 & 7.48 & 29.89 \\
\model{}
& 36.05 & 44.34 & 74.62 & 82.57 & 8.00 & 12.55 & 58.23
& 75.35 & 5.42 & 19.24 \\

\midrule
\grayrow\multicolumn{11}{c}{\textbf{gemini-embedding-2 / Gemini 3.1 Flash-Lite}} \\
\midrule
Oracle
& - & - & - & - & - & - & -
& 91.26 & 2.46 & 6.28 \\
Full KB
& - & - & - & - & - & - & -
& 73.07 & 8.71 & 18.22 \\
Sparse
& 19.77 & 26.52 & 51.45 & 62.61 & 2.35 & 4.90 & 37.82
& 67.44 & 8.00 & 24.56 \\
Dense
& 14.04 & 20.36 & 37.58 & 49.60 & 1.60 & 3.12 & 26.90
& 63.59 & 9.39 & 27.02 \\
\model{}
& 35.37 & 42.29 & 70.50 & 78.03 & 8.67 & 12.15 & 57.48
& 77.44 & 6.49 & 16.07 \\
\bottomrule
\end{tabular}
}
\caption{Main benchmarking results on \dataset{}. All metrics are reported on a percentage scale (\%).}
\label{tab:mainBenchmark}
\end{table*}

\subsection{Evaluation Metrics}
Our evaluation covers two dimensions: retrieval performance and response quality.

\sparagraph{Retrieval Performance.}
For retrieval methods, we measure the quality of the retrieved document set using Recall@$K$, Hit@$K$, Gold@$K$, and MRR. Recall@$K$ measures the fraction of gold documents recovered within the top-$K$ retrieved results. Hit@$K$ measures whether at least one gold document appears in the top-$K$ results. Gold@$K$ measures whether all gold documents are recovered within the top-$K$ results. MRR measures the mean reciprocal rank of the first relevant document across queries. We report results at $K \in \{5, 10\}$.

\sparagraph{Response Quality.}
We evaluate response quality by comparing each model output against the gold rationale, using \texttt{GPT-5.4-mini} as an automatic judge. Since model outputs are free-form natural language rather than structured labels, the judge evaluates both whether the response conveys the correct outcome and whether its rationale aligns with the gold rationale. We adopt a three-level evaluation scheme: \textsc{Pass}, \textsc{Wrong}, and \textsc{Fail}. \textsc{Pass} indicates that the response conveys the correct outcome and that its rationale captures the core reasoning of the gold response. \textsc{Wrong} indicates that the outcome is correct, but the rationale is incomplete, omits key evidence, or includes irrelevant justifications. \textsc{Fail} indicates that the response gives an incorrect outcome, and that its rationale contradicts or substantially diverges from the gold reasoning. We report the \textsc{Pass} rate as the primary metric for response quality.

To verify our automatic judge's reliability and alignment with human judgment, we conduct a human evaluation using Amazon Mechanical Turk.\footnote{\url{https://www.mturk.com/}} We randomly sample 600 instances and collect annotations from three annotators per instance, each of whom makes a binary judgment on whether the assigned label is correct. The final label for each instance is determined by majority voting. The resulting agreement rate between human annotators and the automatic judge is 93.5\%, confirming the reliability of our automatic evaluation. Further details are provided in Appendix~\ref{app:ResponseQualHumanVal}.

\begin{table*}[t]
\centering
\resizebox{0.92\linewidth}{!}{
\begin{tabular}{l c c c c c c c c c}
\toprule
\textbf{Method}
& \multicolumn{3}{c}{\textbf{Overall Quality.}} 
& \multicolumn{3}{c}{\textbf{Conflict Query.}} 
& \multicolumn{3}{c}{\textbf{Non-conflict Query.}} \\
\cmidrule(lr){2-4}\cmidrule(lr){5-7}\cmidrule(lr){8-10}
& \textbf{\textsc{Pass}} & \textbf{\textsc{Wrong}} & \textbf{\textsc{Fail}}
& \textbf{\textsc{Pass}} & \textbf{\textsc{Wrong}} & \textbf{\textsc{Fail}}
& \textbf{\textsc{Pass}} & \textbf{\textsc{Wrong}} & \textbf{\textsc{Fail}} \\
\midrule
\grayrow\multicolumn{10}{c}{\textbf{Qwen3-Embedding-8B / Qwen3-4B-Instruct-2507}} \\
\midrule
Oracle
& 86.89 & 1.69 & 11.42 
& 79.46 & 2.07 & 18.46 
& 94.47 & 1.31 & 4.23 \\
Full KB
& 57.49 & 12.10 & 30.41 
& 30.77 & 14.08 & 55.15 
& 84.76 & 10.07 & 5.16 \\
Sparse
& 62.73 & 8.13 & 29.15 
& 41.80 & 9.63 & 48.57 
& 84.08 & 6.59 & 9.33 \\
Dense
& 62.39 & 8.40 & 29.21 
& 37.78 & 10.85 & 51.37 
& 87.50 & 5.91 & 6.59 \\
\model{}
& 68.82 & 7.60 & 23.58 
& 53.20 & 7.86 & 38.94 
& 84.76 & 7.34 & 7.90 \\

\midrule
\grayrow\multicolumn{10}{c}{\textbf{text-embedding-3-small / GPT-5.4-mini}} \\
\midrule
Oracle
& 87.13 & 2.09 & 10.77 
& 76.54 & 3.23 & 20.23 
& 97.95 & 0.93 & 1.12 \\
Full KB
& 73.10 & 7.29 & 19.61 
& 55.70 & 9.20 & 35.10 
& 90.86 & 5.35 & 3.79 \\
Sparse
& 65.53 & 7.05 & 27.42 
& 40.28 & 8.04 & 51.68 
& 91.29 & 6.03 & 2.67 \\
Dense
& 62.63 & 7.48 & 29.89 
& 33.88 & 8.96 & 57.16 
& 91.98 & 5.97 & 2.05 \\
\model{}
& 75.35 & 5.42 & 19.24 
& 59.17 & 5.91 & 34.92 
& 91.85 & 4.91 & 3.23 \\

\midrule
\grayrow\multicolumn{10}{c}{\textbf{gemini-embedding-2 / Gemini 3.1 Flash-Lite}} \\
\midrule
Oracle
& 91.26 & 2.46 & 6.28 
& 92.63 & 1.46 & 5.91
& 89.86 & 3.48 & 6.65 \\
Full KB
& 73.07 & 8.71 & 18.22
& 71.91 & 9.69 & 18.40
& 74.25 & 7.71 & 18.03 \\
Sparse
& 67.44 & 8.00 & 24.56 
& 57.22 & 11.03 & 31.75
& 77.86 & 4.91 & 17.23 \\
Dense
& 63.59 & 9.39 & 27.02
& 49.85 & 14.02 & 36.14
& 77.61 & 4.66 & 17.72 \\
\model{}
& 77.44 & 6.49 & 16.07 
& 75.93 & 6.09 & 17.98
& 78.98 & 6.90 & 14.12 \\

\bottomrule
\end{tabular}
}
\vspace{-3pt}
\caption{Response quality results by feasibility status in three configurations.}
\vspace{-10pt}
\label{tab:mainBenchQuery}
\end{table*}

\subsection{Experimental Setup}
\sparagraph{Model Configurations.}
For the open-source setting, all agent components are powered by \texttt{Qwen3-4B-Instruct-2507}~\cite{yang2025qwen3} and document embeddings are computed using \texttt{Qwen3-Embedding-8B}~\cite{zhang2025qwen3}.
For the closed-source setting, we use two model configurations: \texttt{gpt-5.4-mini} with \texttt{text-embedding-3-small}\footnote{\url{https://developers.openai.com/api/docs/models/text-embedding-3-small}} for the GPT configuration, and \texttt{Gemini 3.1 Flash-Lite}\footnote{\url{https://ai.google.dev/gemini-api/docs/models/gemini-3.1-flash-lite}} with \texttt{gemini-embedding-2}\footnote{\url{https://ai.google.dev/gemini-api/docs/models/gemini-embedding-2}} for the Gemini configuration.
\model{} is model-agnostic and can be readily adapted to other language and embedding models. 
In particular, the effect of embedding model selection in the open-source setting is analyzed in Appendix~\ref{app:embedding}.

\sparagraph{Method.}
We compare \model{} against the following baselines on \dataset{} to evaluate their effectiveness in conflict-aware reasoning over egocentric KBs. Since Oracle and Full KB methods do not perform retrieval, retrieval metrics are not reported for these settings.

\begin{itemize}
[leftmargin=15pt, itemsep=0pt, topsep=3pt]
\item \textbf{Oracle:} As an upper bound, we provide the model with only the gold documents directly relevant to each query and generate the final response without retrieval.

\item \textbf{Full KB:} We provide the model with the entire KB and generate the final response without retrieval. Since all relevant documents are available in principle, this setting measures how effectively the model can reason over a large, unfiltered context.

\item \textbf{Sparse Retrieval:} We retrieve documents using BM25 over tokenized KB documents based on the original query.

\item \textbf{Dense Retrieval:} We retrieve documents using dense vector search based on the original query.

\end{itemize}

\section{Results}
Table~\ref{tab:mainBenchmark} presents the main results on \dataset{}. Across both settings, the Oracle method consistently achieves the highest \textsc{Pass} rate, establishing a strong upper bound. Despite having access to the entire KB, Full KB achieves only 57.49\% in the open-source setting and about 73\% across the two closed-source settings, both markedly below Oracle. This gap indicates that unfiltered context substantially impairs conflict reasoning, and that the model's reasoning capability plays a critical role beyond mere context availability.

In the open-source setting, sparse and dense retrieval achieve 62.73\% and 62.39\%, respectively. \model{} achieves a 68.82\% \textsc{Pass} rate, outperforming all retrieval-based baselines. This result suggests that targeted evidence selection through multi-hop traversal and conflict-aware filtering enables more reliable conflict reasoning than broader context access. Please see Appendix~\ref{app:detailResults} for detailed results, and Appendix~\ref{app:modelOut} for model examples. Moreover, Appendix~\ref{app:crossmodel} presents an evaluation using \texttt{Gemini 3.1 Flash-Lite} as an independent judge to assess potential self-preference bias and confirm the robustness of our results. 

\section{Analysis}

\subsection{Conflict Status vs. Non-conflict Status}
Table~\ref{tab:mainBenchQuery} reports response quality by feasibility status across all methods and model settings. Conflict queries are generally more challenging than Non-conflict queries, with lower \textsc{Pass} rates across all non-oracle methods. This gap is especially notable in the Qwen and GPT configurations, where it persists even under Oracle setting, implying that conflict resolution remains difficult even when the relevant facts are directly provided.

Across all three configurations, \model{} achieves the highest Conflict \textsc{Pass} rate among non-oracle methods. Compared with the strongest non-oracle baseline in each configuration, it improves Conflict \textsc{Pass} by 11.40, 3.47, and 4.02 percentage points, respectively, while maintaining comparable performance on Non-conflict queries. These results highlight the value of targeted multi-hop traversal and conflict-aware filtering in uncovering evidence of implicit constraints.

\subsection{Evidence Coverage}

\begin{figure}[t]
    \centering
    \includegraphics[width=0.999\columnwidth]{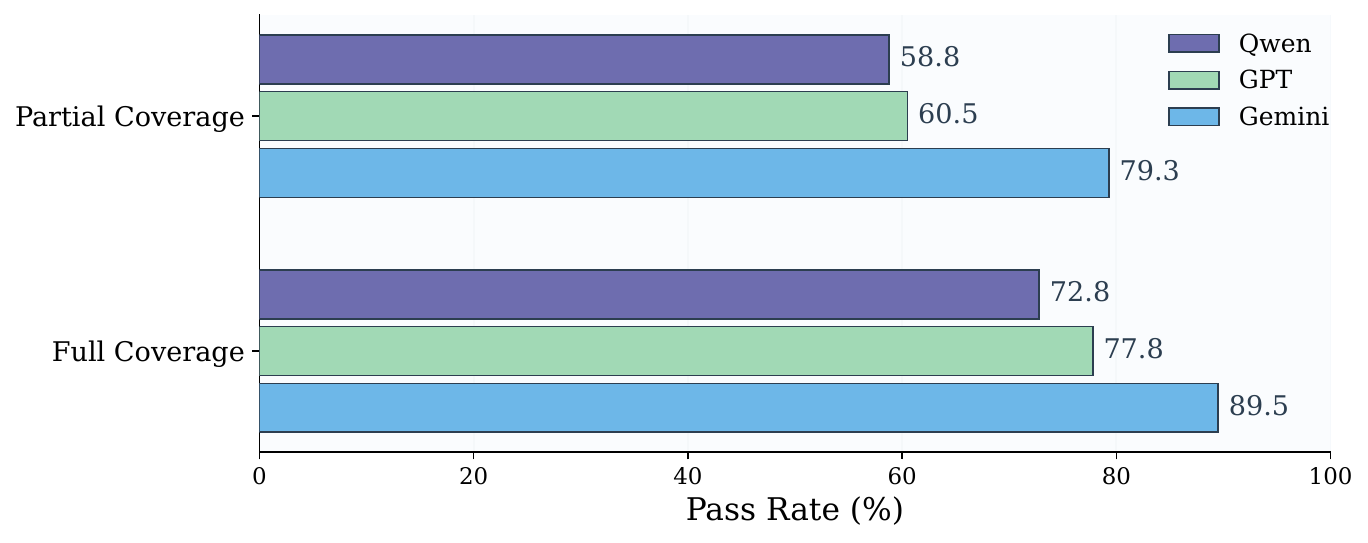}
    \caption{\textsc{Pass} rates by gold evidence coverage for Conflict queries.}
    \label{fig:conflictCoverage}
\end{figure}

Figure~\ref{fig:conflictCoverage} reports \textsc{Pass} rates for Conflict queries under two levels of gold evidence coverage across the three model configurations: partial coverage (Hit@10=1 but Gold@10=0) and full coverage (Gold@10=1). Since the gold contexts in \dataset{} are decomposed into atomic facts distributed across the KB, the evidence required for conflict judgment is rarely contained within a single document. Retrieving only part of these facts already yields a meaningful \textsc{Pass} rate, but full coverage of all gold documents leads to a clear additional gain, suggesting that incomplete evidence is often insufficient for reliable conflict judgment. This result underscores the complexity of surfacing the complete set of conflict-relevant evidence, as the required facts are atomically distributed and not semantically salient from the query alone.

\subsection{Agent Component Ablation}

\begin{table}[t]
\centering
\resizebox{0.9\linewidth}{!}{
\begin{tabular}{l c c c}
\toprule
\textbf{\textsc{Method}} & \textbf{Recall@5} & \textbf{\textsc{Pass}} & \textbf{\textsc{Conflict}}\\
\midrule
w/o Planning & 32.00 & 73.31 & 54.17\\
w/o Traversal & 22.48 & 71.65 & 52.41\\
w/o Selection & 32.61 & 75.19 & 58.87\\
\model{} & 36.05 & 75.35 & 59.17\\
\bottomrule
\end{tabular}
}
\vspace{-3pt}
\caption{An ablation study on \model{}. \textsc{Pass} refers to the overall pass rate, while \textsc{Conflict} refers to the pass rate on Conflict queries only.}
\vspace{-11pt}
\label{tab:ablation}
\end{table}

Table~\ref{tab:ablation} presents an ablation study over the key agent components of \model{}, excluding seed retrieval. To more precisely isolate the contribution of each component, we conduct this analysis under the GPT configuration.

Removing any component consistently degrades performance. The largest drop occurs when multi-hop traversal is removed (71.65\% overall, 52.41\% on Conflict queries), confirming that hop-based expansion is critical for surfacing evidence not directly retrievable from the queries. Removing query planning leads to a comparable decline, while removing evidence selection results in a smaller but still consistent drop.

These results suggest that the three components are mutually reinforcing: query planning retrieves conflict-relevant seeds, traversal expands coverage to indirectly connected evidence, and evidence selection filters the candidate pool to retain decision-critical facts. Please see Appendix~\ref{app:ablation} for detailed experimental setup and results.

\subsection{Comparison with Structured Retrieval Methods}

Table~\ref{tab:comparison} compares \model{} against structured retrieval baselines, GraphRAG~\cite{edge2024local} and HippoRAG 2~\cite{gutierrez2025from}. While \model{} achieves a slightly higher overall \textsc{Pass} rate, the remarkable difference appears on Conflict queries, where it yields a substantially larger improvement over both baselines.

This gap suggests that existing structured retrieval methods can often retrieve documents that are topically relevant to the query, but still fail to recover the full set of conflict-inducing constraints needed for the final decision. Thus, the challenge is not simply multi-hop retrieval, but retrieval guided toward decisive evidence. \model{} addresses this issue through conflict-aware query planning and targeted evidence filtering, enabling it to surface the latent constraints that determine whether a request can be fulfilled.

The methods also differ in their computational profiles. Although HippoRAG 2 offers faster online retrieval once its index is built, \model{} uses no LLM calls during indexing and incurs far fewer calls in total, leading to a lower cold-start cost. This property is particularly beneficial for personalized KBs that may be initialized or refreshed frequently. Further details on the structured retrieval baselines and computational cost measurement are provided in Appendix~\ref{app:comparison} and Appendix~\ref{app:compCost}, respectively.

\begin{table}[t]
\centering
\resizebox{0.9\linewidth}{!}{
\begin{tabular}{l c c c}
\toprule
\textbf{\textsc{Method}} & \textbf{Recall@5} & \textbf{\textsc{Pass}} & \textbf{\textsc{Conflict}}\\
\midrule
HippoRAG 2 & 20.15 & 65.13 & 43.57\\
GraphRAG & - & 58.26 & 34.98\\
\model{} & 26.78 & 68.67 & 54.42\\
\bottomrule
\end{tabular}
}
\vspace{-3pt}
\caption{Comparison of \model{} with structured retrieval methods. We use \texttt{Qwen3-4B-Instruct-2507} as the LLM and \texttt{NV-Embed-v2} as the embedding model for a consistent experimental setting.}
\vspace{-11pt}
\label{tab:comparison}
\end{table}

\section{Conclusion}
In this work, we address the challenge of conflict-aware reasoning over egocentric KBs, where a model must determine whether a user request conflicts with personal context distributed across thousands of atomic facts. To this end, we introduce \dataset{}, a dataset designed to evaluate this capability, and \model{}, a training-free multi-agent retrieval framework that identifies decision-relevant evidence through conflict-aware query planning, hybrid retrieval, multi-hop graph traversal, and conflict-aware evidence selection. Experiments on \dataset{} demonstrate that \model{} outperforms retrieval-based baselines, and our analysis shows that retrieval completeness is particularly critical for Conflict queries, where missing evidence substantially degrades reasoning quality. Our results suggest that substantial room remains for future work on conflict-aware reasoning and retrieval.

\section*{Limitations}
Our benchmark focuses primarily on feasibility judgment rather than full task execution. Although this design isolates the problem of detecting latent conflict signals in user specific context, it does not evaluate whether an assistant can complete complex downstream tasks such as recommendation, scheduling, or planning. Extending the knowledge base with richer persona worlds, actionable entities, and dynamic task environments would enable future benchmarks to assess both conflict awareness and end-to-end personalized task solving. In addition, our method is evaluated without task specific training. While this training-free setup highlights the generality of the proposed framework, future work could train specialized agents for query reformulation, evidence selection, graph traversal, and decision calibration to enhance performance. 

\section*{Ethics Statement}
Our dataset is constructed using synthetic identity profiles and does not contain real personal information. Although fictional, the profiles and knowledge bases are designed to emulate realistic personal contexts and should not be used for privacy-invasive profiling, manipulation, or inappropriate personalization. As the task concerns conflict-aware reasoning, some queries may involve sensitive or potentially unsafe situations. We filter out inherently harmful, unreasonable, or unethical requests, but users of the benchmark should remain cautious when interpreting or extending the data. We use large language models to assist data generation and minor language polishing, while the overall research process is led and conducted primarily by the authors.
\section*{Acknowledgments}
We thank the reviewers and the action editor for their valuable feedback. This work was partly supported by the Institute of Information \& communications Technology Planning \& Evaluation(IITP) grant funded by the Korea government(MSIT) (No.RS-2019-II191906, Artificial Intelligence Graduate School Program(POSTECH), Contribution Rate: 35\%), the Institute of Information \& communications Technology Planning \& Evaluation(IITP) under the Leading Generative AI Human Resources Development (IITP-2026-RS-2026-25546560, Contribution Rate: 35\%) grant funded by the Korea government(MSIT), and the Institute of Information \& Communications Technology Planning \& Evaluation(IITP)-ITRC(Information Technology Research Center) grant funded by the Korea government(MSIT) (IITP-2026-RS-2024-00437866, Contribution Rate: 30\%).

\bibliography{custom}

\clearpage
\appendix

\section{Dataset Details}
\label{app:dataDetail}
Table~\ref{tab:QueryDistri} presents the distribution by situation type. Additional examples are provided in Table~\ref{tab:PerExpan} for persona expansion; Table~\ref{tab:PerSynth1} and Table~\ref{tab:PerSynth2} for profile synthesis; Table~\ref{tab:QueryExam1} for Conflict queries; and Table~\ref{tab:QueryExam2} for Non-conflict queries.

\begin{table}[h]
\centering
\resizebox{0.9\linewidth}{!}{
\begin{tabular}{l c c c}
\toprule
\textbf{Type} & \textbf{Conflict} & \textbf{Non-conflict} & \textbf{Total}\\
\midrule
Temporal & 442 & 595 & 1,037\\
Personal & 661 & 470 & 1,131\\
State & 538 & 543 & 1,081\\
\midrule
Total & 1,641 & 1,608 &3,249\\
\bottomrule
\end{tabular}
}
\caption{Query distribution of \dataset{}.}
\label{tab:QueryDistri}
\end{table}

\section{Implementation Details}
\label{app:implDetail}
\model{} consists of multiple agent components. Since \model{} is training-free, we use the models without additional fine-tuning. The key hyperparameters for each component, including the offline index construction, are as follows:

\begin{itemize}
[leftmargin=15pt, itemsep=0pt, topsep=3pt]

\item \textbf{Indexing:} We encode all KB documents using the embedding model and construct a k-nearest neighbor (k-NN) document graph based on cosine similarity. We set the number of neighbors per document to $k=10$.

\item \textbf{Query Planning:} The conflict planner generates up to $3$ conflict dimensions per query, which are passed to the query generator to produce up to $3$ counter queries alongside the original query.

\item \textbf{Hybrid Retrieval:} For each query view, we retrieve the top-$K$ documents from both dense and sparse retrievers, where $K=10$. The retrieved results are merged using Weighted Reciprocal Rank Fusion (WRRF) with $k=60$. Counter query results are assigned a weight of $1.2$, while original-query results are assigned a weight of $1.0$. The top-$N_{\text{seed}}=20$ fused documents are passed to the pre-hop filter as candidate seed documents.

\item \textbf{Multi-Hop Graph Traversal:} The top-$N_{\text{filter}}=10$ seed documents selected by pre-hop filter agent serve as entry points for BFS-based traversal up to a maximum depth of $H=5$ hops, expanding $M=3$ neighboring documents at each hop.

\item \textbf{Evidence Selection:} After traversal, the post-hop filter agent selects the top-$N_{\text{final}}=10$ documents from the collected document pool as the final evidence set passed to the answer generator.

\end{itemize}

We use FAISS~\cite{douze2025faiss} for dense index construction and retrieval, and BM25 for sparse retrieval. For the open-source setting, all agent models are served via vLLM~\cite{kwon2023efficient}, and all experiments are conducted on NVIDIA RTX A6000 4EA GPUs.

We examine the sensitivity of retrieval and traversal hyperparameters in the GPT setting, including traversal depth ($H$), neighbors per document ($M$), fusion weights, top-$K$, and top-$N$. As shown in Table~\ref{tab:hyperparameter_sensitivity} and Table~\ref{tab:hyperparameter_sensitivity_response}, the configuration used in our experiments achieves the strongest results, while performance remains stable across nearby configurations.

\section{Human Evaluation Details}
\label{app:humanEval}
We conduct human evaluations to verify that (1) the feasibility status assigned to each query is supported by the provided evidence, and (2) our automatic judge is aligned with human judgment and produces reliable evaluations. To ensure annotation quality, we restrict participation to workers with more than 10,000 approved HITs and a HIT approval rate above 98\%. Each evaluation instance is assessed by three annotators, and the final label is determined by majority vote.

\subsection{Feasibility Status}
\label{app:feasibilityHumanVal}

\begin{table}[h]
\centering
\resizebox{0.9\linewidth}{!}{
\begin{tabular}{l c c c}
\toprule
\textbf{Type}
& \textbf{GPT--Human}
& \textbf{Inter-Annotator} \\
\midrule
Overall      & 93.3\% & 86.7\% \\
Conflict     & 88.3\% & 80.0\% \\
Non-conflict & 98.3\% & 93.3\% \\
\bottomrule
\end{tabular}
}
\caption{Agreement results from human evaluation.}
\label{tab:human_eval_feasibility}
\end{table}

Although the Conflict and Non-conflict feasibility statuses are assigned during dataset construction, we further evaluate their alignment with human judgments. For each of the three situation types (temporal, personal, and state), we randomly sample 20 Conflict and 20 Non-conflict queries, resulting in 120 queries in total. Annotators assess each query and its corresponding gold facts to determine whether the request is feasible and non-problematic.

As shown in Table~\ref{tab:human_eval_feasibility}, the feasibility status assignments achieve 93.3\% overall agreement with human annotations, with an overall inter-annotator agreement of 86.7\%. Agreement reaches 98.3\% for Non-conflict queries and 88.3\% for Conflict queries.

The remaining disagreements occur primarily among Conflict queries. Manual inspection of these cases suggests that human annotators occasionally adopt a more permissive interpretation of feasibility, whereas our construction pipeline applies the feasibility criteria more strictly. Nevertheless, the high overall agreement demonstrates that the feasibility statuses are well aligned with human judgments.

\subsection{Response Quality}
\label{app:ResponseQualHumanVal}

To validate the automatic judgments of response quality, we sample 200 instances per judge label (\textsc{Pass}, \textsc{Wrong}, and \textsc{Fail}) from the evaluation results of \model{} in the open-source setting, resulting in a total of 600 instances. Within each label category, instances are sampled randomly while maintaining a balanced ratio of Conflict and Non-conflict queries.

Each annotator makes a binary judgment on whether the assigned label is correct.
Overall, 72.5\% of instances achieve full agreement among all three annotators, while the remaining 27.5\% show 2 out of 3 agreement, indicating high inter-annotator consistency.

\begin{figure*}
  \centering
  \includegraphics[width=1.95\columnwidth]{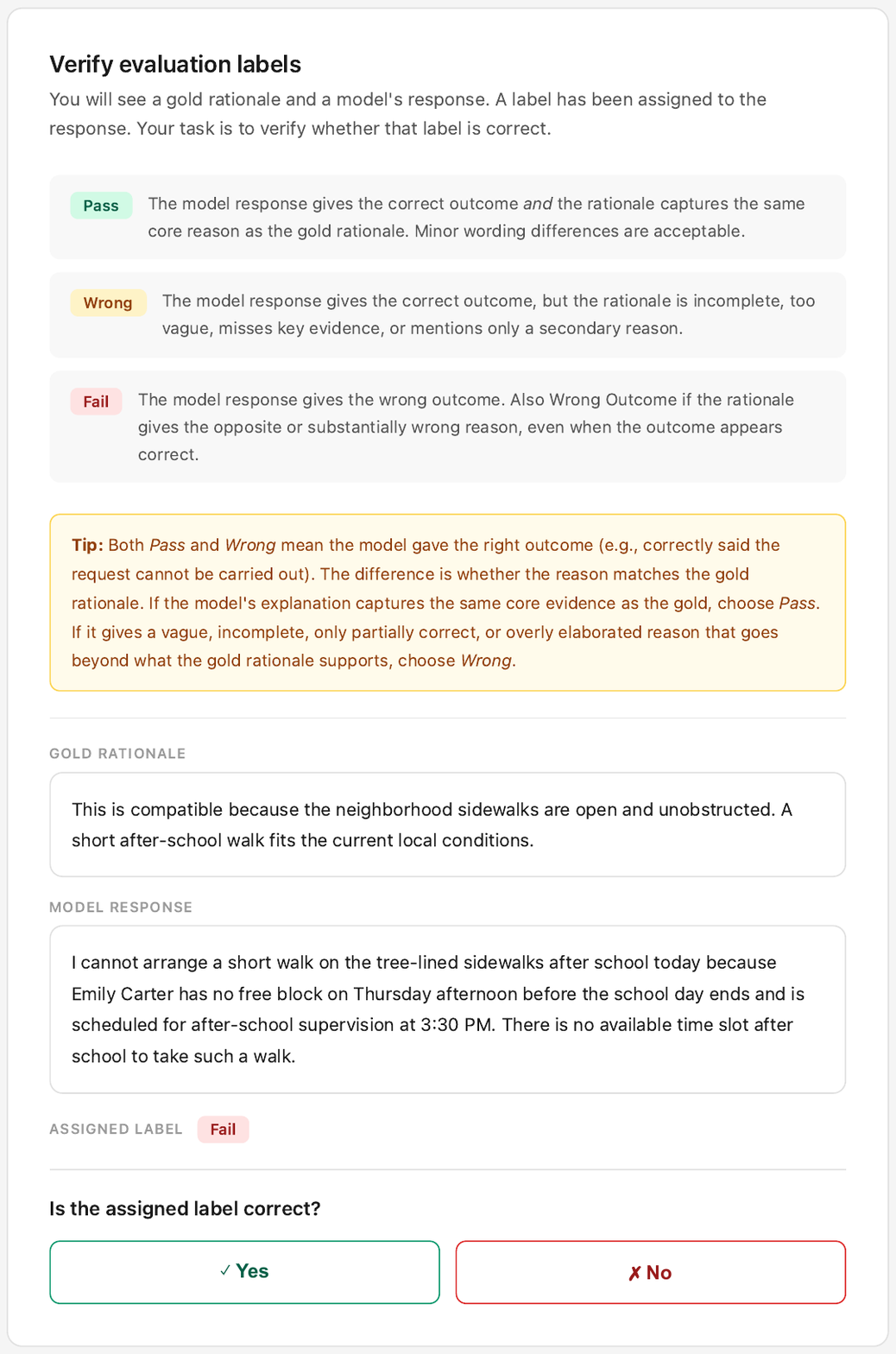}
  \caption{Annotation interface for human evaluation.}
    \label{fig:humanEval}
\end{figure*}

The overall agreement rate between human annotators and the automatic judge is 93.5\%, confirming that our automatic judge is well aligned with human judgment. By label, the agreement rates are 97.0\% for \textsc{Pass}, 90.5\% for \textsc{Wrong}, and 93.0\% for \textsc{Fail}. The relatively lower agreement for \textsc{Wrong} is expected, as annotators may differ in how strictly they assess whether a model-generated rationale matches the gold rationale. Since our judge is designed to evaluate rationale alignment strictly against the gold rationale, this behavior is consistent with the intended evaluation criteria and does not undermine the overall reliability of the judge. Please see Figure~\ref{fig:humanEval} for the human evaluation interface.

\section{Further Analysis}
\label{app:analysis}

\subsection{Embedding Model Selection}
\label{app:embedding}
For the open-source setting, we select \texttt{Qwen3-4B-Instruct-2507}~\cite{yang2025qwen3} as the agent model, due to its strong efficiency–performance trade-off and broad adoption in open-source agentic systems. For the embedding model, we conduct experiments with three widely used models to analyze the effect of embedding model choice on retrieval and overall performance. Specifically, we compare \texttt{Qwen3-Embedding-8B}~\cite{zhang2025qwen3}, \texttt{BGE-M3}~\cite{chen-etal-2024-m3}, and \texttt{NV-Embed-v2}~\cite{lee2025nvembed}, which are known for strong performance in dense retrieval.

Retrieval and overall performance are reported in Table~\ref{tab:embed1}, while performance by feasibility status is shown in Table~\ref{tab:embed2}. Interestingly, all three embedding models yield very similar performance across metrics. This suggests that, in our setting, the quality of conflict-aware reasoning and agentic filtering also plays a significant role in the final outcome, alongside the choice of embedding model. Based on these results, we select \texttt{Qwen3-Embedding-8B} as our embedding model for the open-source setting, as it achieves the best overall performance and belongs to the same model family as the agent model.
    
\subsection{Detailed Results}
\label{app:detailResults}
\begin{figure*}[t]
    \centering
    \includegraphics[width=1.95\columnwidth]{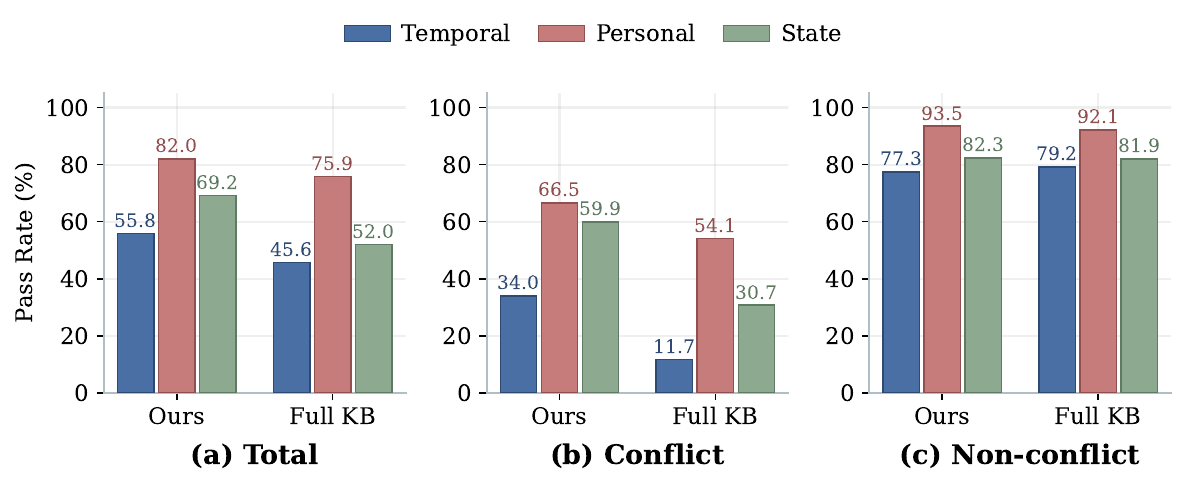}
    \caption{\textsc{Pass} rates by situation type and feasibility status in the Qwen setting.}
    \label{fig:typewise_qwen}

    \vspace{0.8em}
    
    \includegraphics[width=1.95\columnwidth]{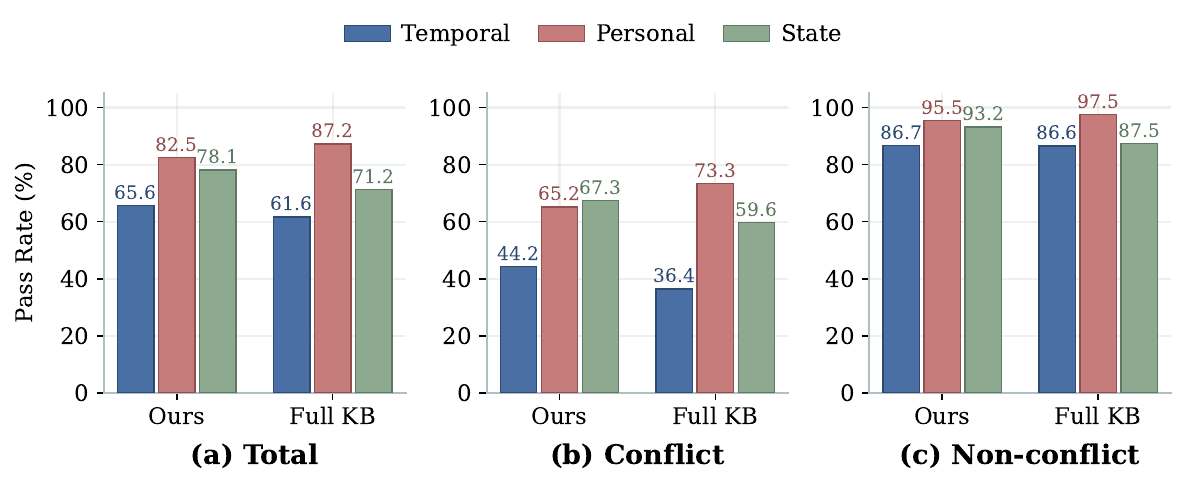}
    \caption{\textsc{Pass} rates by situation type and feasibility status in the GPT setting.}
    \label{fig:typewise_gpt}

    \vspace{0.8em}

    \includegraphics[width=1.95\columnwidth]{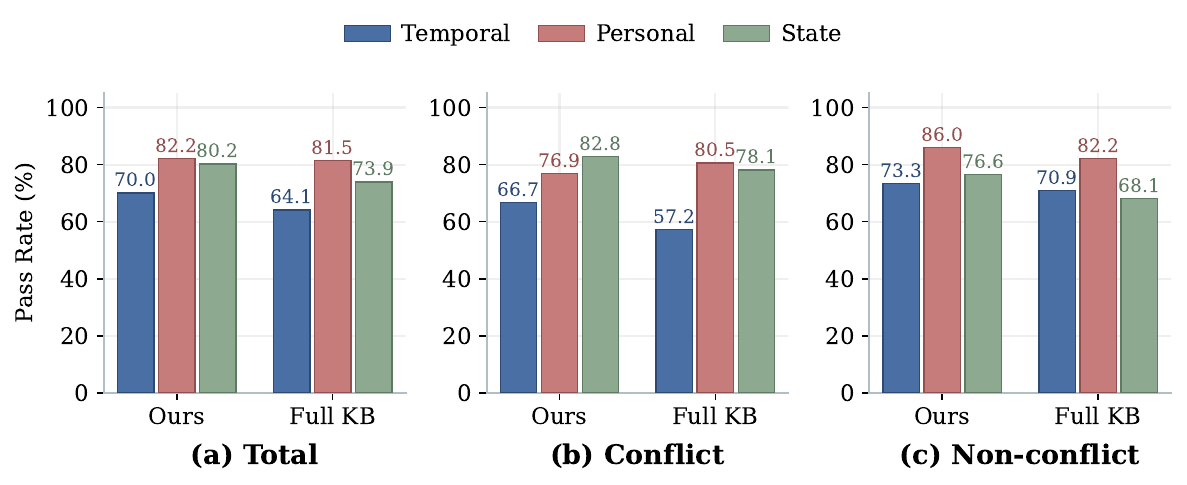}
    \caption{\textsc{Pass} rates by situation type and feasibility status in the Gemini setting.}
    \label{fig:typewise_gemini}
\end{figure*}

Figure~\ref{fig:typewise_qwen}, Figure~\ref{fig:typewise_gpt}, and Figure~\ref{fig:typewise_gemini} show \textsc{Pass} rates by situation type across the Qwen, GPT, and Gemini configurations. Across all three settings, performance varies across situation types, with temporal queries consistently showing the lowest \textsc{Pass} rates. We believe this is because temporal conflicts often involve routines, recurring schedules, or time-related commitments that are connected to one another. To resolve these conflicts, the model must reason over multiple temporally related facts instead of relying on a single decisive piece of evidence. This makes evidence aggregation more difficult, since the relevant information is spread across several facts that may not seem important when viewed individually from the original query alone.

\subsection{Model Outputs}
\label{app:modelOut}
Please refer to Table~\ref{tab:ModelExamTemp} for temporal-type query response examples, Table~\ref{tab:ModelExamPers} for personal-type query response examples, and Table~\ref{tab:ModelExamState} for state-type query response examples.

\subsection{Cross-Model Judge Evaluation}
\label{app:crossmodel}

To investigate the robustness of our evaluation to the choice of judge model, we evaluate all \model{} responses generated under the open-source setting using \texttt{Gemini 3.1 Flash-Lite} as an alternative judge model.
As shown in Table~\ref{tab:judge_robustness}, \texttt{Gemini 3.1 Flash-Lite} and \texttt{GPT-5.4-mini} judges exhibit an 86.40\% agreement rate across \textsc{Pass}, \textsc{Wrong}, and \textsc{Fail}. Especially, the two judges show virtually identical \textsc{Fail} rates across overall (23.61\% vs. 23.58\%), Conflict (38.09\% vs. 38.94\%), and Non-conflict queries (8.83\% vs. 7.90\%). Disagreements primarily emerge between the \textsc{Pass} and \textsc{Wrong} categories, where \texttt{Gemini 3.1 Flash-Lite} tends to label Conflict responses as \textsc{Wrong} while assigning Non-conflict responses to \textsc{Pass}. This pattern implies that both judges reliably agree on overall failure cases but apply slightly different criteria when evaluating rationale quality.

\begin{table}[t]
\centering
\resizebox{0.99\linewidth}{!} {
\begin{tabular}{l c c c c c c}
    \toprule
    & \multicolumn{3}{c}{\textbf{GPT-5.4-mini}}
    & \multicolumn{3}{c}{\textbf{Gemini 3.1 Flash-Lite}} \\
    \cmidrule(lr){2-4} \cmidrule(lr){5-7}
    \textbf{Split}
    & \textsc{Pass} & \textsc{Wrong} & \textsc{Fail}
    & \textsc{Pass} & \textsc{Wrong} & \textsc{Fail} \\
    \midrule
    Overall
    & 68.82 & 7.60 & 23.58
    & 67.25 & 9.14 & 23.61 \\
    Conflict
    & 53.20 & 7.86 & 38.94
    & 45.58 & 16.33 & 38.09 \\
    Non-conflict
    & 84.76 & 7.34 & 7.90
    & 89.37 & 1.80 & 8.83 \\
    \bottomrule
\end{tabular}
}
\caption{Evaluation results with different judge models.}
\label{tab:judge_robustness}
\end{table}

\subsection{Ablation Settings}
\label{app:ablation}
We conduct an ablation study over the three core agent components of \model{}. The detailed configuration of each ablation setting is as follows:

\begin{itemize}
[leftmargin=15pt, itemsep=0pt, topsep=3pt]

\item \textbf{w/o Planning:} This setting removes the conflict-aware query planning stage entirely. As a result, no counter queries are generated, and retrieval is performed using only the original query.

\item \textbf{w/o Traversal:} This setting removes the multi-hop graph traversal stage, including all hop-based expansion. Candidate documents obtained through hybrid retrieval and fusion are passed directly to the answer generator without traversal.

\item \textbf{w/o Selection:} This setting removes the post-hop evidence selection stage. All documents collected after traversal are passed directly to the answer generator without further filtering.

\end{itemize}

The full ablation results are reported in Table~\ref{tab:fullAbla1}, with results for feasibility status provided in Table~\ref{tab:fullAbla2}.

\subsection{Comparison with Other Methods}
\label{app:comparison}
We compare our method with two structure-aware RAG baselines, HippoRAG 2~\cite{gutierrez2025from} and GraphRAG~\cite{edge2024local}. To ensure a controlled comparison, all methods use \texttt{Qwen3-4B-Instruct-2507} as the LLM backbone wherever LLM inference is required, including graph construction, retrieval, and answer generation. We choose \texttt{NV-Embed-v2} as the common embedding backbone since it can be integrated into our implementations of all three compared methods and was among the strongest publicly available retrieval embeddings~\cite{lee2025nvembed,gutierrez2025from}.

For each egocentric KB instance, we build a separate index or graph and run the corresponding queries independently. Because our task requires judging whether a user request is feasible with respect to an egocentric knowledge base and a predefined reference date, the raw output format of general-purpose RAG systems is not directly comparable. We therefore use the same task-specific answer generator across all methods, which receives the retrieved or generated context and produces whether the request can be fulfilled.

For HippoRAG 2, we use its standard graph-based retrieval pipeline, which performs retrieval over a knowledge graph using Personalized PageRank. While preserving the default retrieval setting, we retrieve the top-$10$ passages, matching the final evidence budget used by our method. Since HippoRAG 2 returns ranked passage candidates aligned with the original corpus, we report both retrieval metrics and model response quality.

For GraphRAG, we use its standard indexing workflow and local search method, which constructs a knowledge graph, detects community structure, and uses graph-derived summaries during query answering. We preserve the default pipeline structure and prompts, and only modify the model configuration to route LLM calls to the controlled backbone model. 

As GraphRAG local search directly produces a synthesized answer from its internal graph and summary representation, we treat the local search output as intermediate context and pass it to the shared answer generator to obtain the final evaluated response. Since GraphRAG local search does not directly return a ranked document list aligned with our data, we report GraphRAG only in the answer quality comparison.

To prevent unbounded execution for unusually long local search calls, we impose a 300-second timeout per query and treat timed-out queries as having no usable returned context. In our runs, completed GraphRAG local search calls with recorded runtimes took 16.3 seconds on average, with a median of 14.0 seconds, while 94 out of 3,249 queries (2.89\%) exceeded the timeout budget. The full comparison results are reported in Table~\ref{tab:fullComp1}, while the results for the feasibility status are provided in Table~\ref{tab:fullComp2}.

\subsection{Computational Cost Analysis}
\label{app:compCost}

To examine the computational cost of \model{} and other structure-aware methods, we measure latency and the number of LLM calls. Since all methods use the same downstream answer generation and evaluation modules, we focus on offline indexing and online retrieval. Table~\ref{tab:computational_cost} reports the results on a randomly sampled instance containing 2,056 facts and 18 queries.

\model{} takes 80.78 seconds for offline indexing and 7.46 seconds per query for online retrieval, yielding a cold-start total of 215.10 seconds. HippoRAG 2 takes 290.36 seconds for indexing and 0.80 seconds per query for retrieval, with a cold-start total of 310.99 seconds, while GraphRAG takes 489.37 seconds, 14.00 seconds, and 741.44 seconds, respectively. 

We also count the LLM calls incurred during indexing and retrieval. HippoRAG 2 makes 4,096 calls during indexing and 18 during retrieval. In contrast, \model{} makes no LLM calls during indexing and 72 calls during retrieval across the 18 queries. Its four calls per query correspond to the Conflict Planner, Multi-View Generator, Pre-hop Filter Agent, and Post-hop Filter Agent. These results show that HippoRAG 2 shifts more computation to offline indexing in exchange for faster subsequent retrieval, while \model{} maintains a lighter indexing stage and lower cold-start cost. 

\begin{table}[t]
\centering
\resizebox{0.95\linewidth}{!}{
\begin{tabular}{m{0.99\linewidth}}
\toprule
\textbf{Original Persona}\\
\midrule
- I am a fashion model and actor.\\
- I do not sing very well.\\
- I grew up in california.\\
- I love to surf and skate.\\
- I will be moving to new york city next year.\\
\midrule
\textbf{Expanded Persona}\\
\midrule
The persona is a fashion model and actor who grew up in California and has a relaxed, outdoors-oriented lifestyle shaped by time near the coast. They enjoy surfing and skateboarding, and those interests continue to play a regular part in their routine, along with the pace and practical demands of work in fashion and screen performance. Their background gives them a comfortable familiarity with casual, sun-driven environments, and they tend to move easily between creative settings and everyday physical activity. They do not sing very well, and they are generally more associated with visual and performance-based work than with music. They are preparing to move to New York City next year, where they will likely adjust to a denser, more fast-moving urban environment while continuing their professional focus.\\
\bottomrule
\end{tabular}
}
\caption{Example of persona expansion.}
\label{tab:PerExpan}
\end{table}

\begin{table*}
\centering
\resizebox{0.95\linewidth}{!}{
\begin{tabular}{m{0.99\linewidth}}
\toprule
\textbf{- Demographic}\\
name: Evan Mateo Reyes\\
age: 29\\
gender: Male\\
marital\_status: In a relationship\\
family: Lives with his girlfriend and a rescue terrier mix; maintains close contact with his parents and younger sister in California\\
\\
\textbf{- Geographic}\\
country: United States\\
city: Los Angeles\\
neighborhood: Venice\\
housing: 2-bedroom loft apartment with exposed brick, floor-to-ceiling windows, and a secured surfboard rack in the garage\\
environment: Flat coastal neighborhood with heavy morning marine layer, bike lanes, and quick access to the beach; usually breezy, sunlit, and active\\
mobility: Street parking is limited, rideshare pickup is easy, there is a Metro bus stop within a 5-minute walk, and a 24-hour convenience store plus a small surf shop are downstairs\\
\\
\textbf{- Socio-Economic}\\
occupation: Fashion model and actor\\
organization: Freelance representation through Ford Models and Atlas Talent Agency\\
income\_tier: High-income\\
financial\_status: Comfortably paid but variable month-to-month due to irregular booking cycles; saving aggressively for a New York City move and keeping a separate tax reserve account\\
education: Completed two years of a BA in Film and Media Studies at University of California, Santa Barbara before pausing for modeling work\\
\\
\textbf{- Health \& Physical}\\
conditions: Seasonal allergic rhinitis, Mild exercise-induced asthma\\
limitations: Occasional knee soreness after long skate sessions or runway travel days, Does not have strong singing ability and avoids vocal performance auditions, Needs to manage asthma triggers during cold-weather outdoor shoots\\
devices: ResMed AirMini portable CPAP used occasionally during travel after sleep-disrupting shoot schedules, Theragun Mini massage device for post-surf recovery, Custom-form orthotic insoles for long casting days\\
medications:\\
- Albuterol inhaler (Relief of mild exercise-induced asthma symptoms during workouts or surf sessions)\\
- Loratadine 10 mg (Seasonal allergy relief)\\
dietary: Avoids shellfish because of a mild allergy history, Limits heavy dairy before early call times to reduce bloating and congestion\\
\\
\textbf{- Tech \& Devices}\\
os: iOS\\
devices: iPhone 15 Pro, 14-inch MacBook Pro with M3 chip, Apple Watch Series 9, Sony WH-1000XM5 headphones\\
apps: Instagram, Notion, Spotify\\
subscriptions: iCloud+ 2 TB, Spotify Premium, Adobe Creative Cloud\\
\bottomrule
\end{tabular}
}
\caption{Example of profile synthesis (Part 1).}
\label{tab:PerSynth1}
\end{table*}

\begin{table*}
\centering
\resizebox{0.95\linewidth}{!}{
\begin{tabular}{m{0.99\linewidth}}
\toprule
\textbf{- Lifestyle}\\
vehicle: 2023 Mercedes-Benz GLC 300 with roof racks for surfboards and a beach towel kit in the cargo area\\
hobbies: Surfing with a 6'2'' shortboard at Malibu and Topanga, Skateboarding on a Santa Cruz setup with independent trucks and soft wheels, Beach workouts and mobility drills using resistance bands and a balance board\\
weekly\_routine:\\
- Monday 7:00 AM surf session followed by a protein-heavy breakfast\\
- Wednesday afternoon skate cruise and casting calls in West Hollywood\\
- Friday early gym session with a trainer focused on mobility and core stability\\
- Sunday long beach walk, laundry, and schedule planning for the coming week\\
\\
\textbf{- Values \& Views}\\
food\_likes: Cold brew coffee, Açaí bowls, Tacos al pastor, Coconut water\\
food\_dislikes: Overly sweet protein shakes, Heavy cream sauces, Shellfish, Very loud nightlife scenes\\
ethics: Prefers durable, low-waste basics, buys sunscreen and skincare with reef-safe ingredients when possible, and favors second-hand vintage outerwear for casual off-duty looks\\
politics: Generally socially liberal and environmentally conscious, supportive of coastal conservation, fair labor in fashion, and broader mental-health awareness in creative industries\\
\\
\textbf{- Relationship}\\
name: Maya Chen\\
age: 28\\
gender: Female\\
city: New York, Manhattan\\
occupation: Model and aspiring actress\\
relation: professional acquaintance\\
traits: disciplined, style-conscious, ambitious, city-savvy, likes singing casually\\
conflicts:\\
- [schedule\_conflict] She often has early call times and last-minute castings that can change her availability the same day.\\
- [preference\_conflict] She is comfortable singing casually and will sometimes suggest karaoke or music-heavy nights, which Evan usually avoids.\\
shared\_context:\\
- They met through overlapping fashion work and have compared casting notes before.\\
- She is based in New York, where Evan expects to move next year.\\
- They occasionally text about auditions, fittings, and agency contacts.\\
\bottomrule
\end{tabular}
}
\caption{Example of profile synthesis (Part 2).}
\label{tab:PerSynth2}
\end{table*}

\begin{table*}
\centering
\resizebox{0.95\linewidth}{!}{
\begin{tabular}{m{0.99\linewidth}}
\toprule
\textbf{Situation type:} Temporal\\
\textbf{Query:} Book a 6:30 AM rideshare tomorrow for a beach workout, then take me to a casting in West Hollywood that starts at 11:00 AM. I also need a stop back at my loft for coffee and a quick outfit change before the casting.\\
\textbf{Context:} Tomorrow morning Evan has a beach workout planned that would take about 90 minutes, plus about 20 minutes to get back to his Venice loft and another 25 minutes to change and grab coffee. The drive from Venice to West Hollywood typically takes about 40 minutes without traffic, and the casting starts at 11:00 AM.\\
\textbf{Judgment:} This conflicts with the available time because the workout, loft stop, and cross-town drive add up to more time than the morning allows before the 11:00 AM casting. Even with a 6:30 AM rideshare, the schedule is too compressed to complete all three steps on time.\\
\textbf{Gold docs:}\\
- Evan Mateo Reyes has a beach workout planned for 2025-01-12 morning that takes about 90 minutes.\\
- Evan Mateo Reyes needs about 20 minutes to get from the beach back to his Venice loft on 2025-01-12 morning.\\
- Evan Mateo Reyes needs about 25 minutes to change clothes and get coffee at his Venice loft on 2025-01-12 morning.\\
- The drive from Venice to West Hollywood typically takes about 40 minutes without traffic.\\
- A casting in West Hollywood starts at 11:00 AM on 2025-01-12.\\
\midrule
\textbf{Situation type:} Personal\\
\textbf{Query:} Can you book me and Maya Chen a dinner reservation at a very loud nightclub restaurant this Saturday night, ideally near the speakers?\\
\textbf{Context:} Evan Mateo Reyes dislikes very loud nightlife scenes and generally avoids them. Maya Chen sometimes suggests music-heavy nights, but the decisive constraint here is Evan's own aversion to loud nightlife.\\
\textbf{Judgment:} This conflicts with Evan's established dislike of very loud nightlife. Reserving a table near the speakers would deliberately place him in the kind of environment he avoids.\\
\textbf{Gold docs:}\\
- Evan Mateo Reyes dislikes very loud nightlife scenes.\\
- Evan Mateo Reyes generally avoids very loud nightlife scenes.\\
- Maya Chen sometimes suggests music-heavy nights.\\
\midrule
\textbf{Situation type:} State\\
\textbf{Query:} Can you arrange a surfboard pickup from my loft garage and have it delivered to the downtown set before sunrise? If the pickup needs garage access, use the building code I have on hand.\\
\textbf{Context:} Evan Mateo Reyes keeps his surfboard in a secured rack inside the garage of his Venice loft. The building's garage is only accessible to residents during staffed daytime hours, and the code he has does not override that access rule. Before-sunrise pickups cannot be completed from that garage.\\
\textbf{Judgment:} This conflicts with the building's access state: the garage is not available for pickup before sunrise, and the code Evan has does not bypass the daytime-only access rule. Because the surfboard is stored in that secured garage, the requested handoff cannot be carried out as timed.\\
\textbf{Gold docs:}\\
- Evan Mateo Reyes keeps his surfboard in a secured rack inside the garage of his Venice loft.\\
- The garage of Evan Mateo Reyes's Venice loft is accessible to residents only during staffed daytime hours.\\
- The building code that Evan Mateo Reyes has does not override the garage's resident-only daytime access rule.\\
- Before-sunrise pickups cannot be completed from Evan Mateo Reyes's garage.\\
\bottomrule
\end{tabular}
}
\caption{Examples of Conflict queries on \dataset{}.}
\label{tab:QueryExam1}
\end{table*}

\begin{table*}
\centering
\resizebox{0.95\linewidth}{!}{
\begin{tabular}{m{0.99\linewidth}}
\toprule
\textbf{Situation type:} Temporal\\
\textbf{Query:} Set up a Friday morning gym session with my trainer and add a reminder to bring my Theragun Mini afterward. Keep the core and mobility focus in the plan.\\
\textbf{Context:} Evan Mateo Reyes has a Friday early gym session with a trainer focused on mobility and core stability. He keeps a Theragun Mini for post-surf and post-workout recovery, and he uses it regularly after strenuous sessions. His exercise routine already centers on recovery and mobility work.\\
\textbf{Judgment:} This remains compatible because it aligns with an existing Friday training commitment and with his recovery tools. The request fits the timing and the physical focus already described in his routine.\\
\textbf{Gold docs:}\\
- Evan Mateo Reyes has a Friday early gym session with a trainer.\\
- Evan Mateo Reyes's Friday gym session with a trainer focuses on mobility and core stability.\\
- Evan Mateo Reyes keeps a Theragun Mini for post-surf and post-workout recovery.\\
- Evan Mateo Reyes regularly uses a Theragun Mini after strenuous sessions.\\
\midrule
\textbf{Situation type:} Personal\\
\textbf{Query:} Book a low-key outdoor dinner for me and Maya Chen after our agency calls. Pick a place that does not center on karaoke or loud music.\\
\textbf{Context:} Evan Mateo Reyes occasionally texts with Maya Chen about auditions, fittings, and agency contacts. Maya sometimes suggests karaoke or music-heavy nights, but Evan usually avoids loud nightlife scenes and prefers relaxed, outdoor-oriented settings.\\
\textbf{Judgment:} This is compatible because the request explicitly asks for a quiet outdoor dinner and avoids karaoke or loud music, which are the settings Evan tends to steer clear of. The plan fits their ongoing professional contact and does not conflict with any stated personal constraint.\\
\textbf{Gold docs:}\\
- Evan Mateo Reyes occasionally texts with Maya Chen about auditions, fittings, and agency contacts.\\
- Maya Chen sometimes suggests karaoke nights or music-heavy nights.\\
- Evan Mateo Reyes usually avoids loud nightlife scenes.\\
- Evan Mateo Reyes prefers relaxed, outdoor-oriented settings.\\
\midrule
\textbf{Situation type:} State\\
\textbf{Query:} Book a studio fitting in West Hollywood for next Wednesday afternoon and make sure the plan uses rideshare from Venice. Add a note that parking is not required.\\
\textbf{Context:} Evan Mateo Reyes lives in Venice, where rideshare pickup is easy and street parking is limited. He regularly handles work appointments in West Hollywood, and a rideshare-based trip fits the local transportation setup.\\
\textbf{Judgment:} This is compatible because Venice supports easy rideshare pickup and the request does not depend on street parking. The studio fitting can be handled in a way that matches the neighborhood’s transportation conditions.\\
\textbf{Gold docs:}\\
- Evan Mateo Reyes lives in Venice.\\
- Evan Mateo Reyes can usually get rideshare pickup easily in Venice.\\
- Street parking in Venice is limited.\\
Evan Mateo Reyes regularly handles work appointments in West Hollywood.\\
\bottomrule
\end{tabular}
}
\caption{Examples of Non-conflict queries on \dataset{}.}
\label{tab:QueryExam2}
\end{table*}

\begin{table*}[t]
\centering
\resizebox{0.97\linewidth}{!}{
\begin{tabular}{l c c c c c c c c c c}
\toprule
\textbf{Setting}
& \multicolumn{7}{c}{\textbf{Retrieval Performance.}} 
& \multicolumn{3}{c}{\textbf{Response Quality.}} \\
\cmidrule(lr){2-8}\cmidrule(lr){9-11}
& \textbf{Recall@5}
& \textbf{Recall@10}
& \textbf{Hit@5}
& \textbf{Hit@10}
& \textbf{Gold@5}
& \textbf{Gold@10}
& \textbf{MRR}
& \textbf{\textsc{Pass}}
& \textbf{\textsc{Wrong}}
& \textbf{\textsc{Fail}} \\
\midrule

\rowcolor{gray!15}
\multicolumn{11}{l}{\textbf{Default}} \\

$H{=}5,\ M{=}3,\ \mathbf{w}{=}[1.0,1.2],\ K{=}20,\ N{=}10$
& \textbf{36.05}
& \textbf{44.34}
& \textbf{74.62}
& \textbf{82.57}
& \textbf{8.00}
& \textbf{12.55}
& \textbf{58.23}
& \textbf{75.35}
& \textbf{5.42}
& \textbf{19.24} \\

\rowcolor{gray!15}
\multicolumn{11}{l}{\textbf{Graph Traversal}} \\

$H{=}2,\ M{=}3$
& 34.27
& 41.18
& 72.12
& 79.54
& 8.00
& 11.03
& 56.35
& 73.84
& 5.79
& 20.38 \\

$H{=}8,\ M{=}3$
& 34.39
& 41.72
& 73.55
& 80.72
& 7.49
& 10.96
& 56.36
& 73.99
& 5.69
& 20.31 \\

$H{=}5,\ M{=}1$
& 33.08
& 38.83
& 70.69
& 77.46
& 6.91
& 9.64
& 55.18
& 73.22
& 6.31
& 20.47 \\

$H{=}5,\ M{=}5$
& 34.62
& 42.60
& 74.35
& 82.26
& 7.00
& 11.03
& 57.35
& 73.65
& 6.49
& 19.85 \\

\rowcolor{gray!15}
\multicolumn{11}{l}{\textbf{Fusion Weights}} \\

$\mathbf{w}{=}[1.0,0.8]$
& 34.12
& 41.59
& 72.68
& 80.13
& 7.78
& 11.18
& 55.47
& 73.47
& 6.37
& 20.16 \\

$\mathbf{w}{=}[1.0,1.0]$
& 34.50
& 41.73
& 72.49
& 79.78
& 7.40
& 10.79
& 56.28
& 73.28
& 6.56
& 20.16 \\

$\mathbf{w}{=}[1.0,1.5]$
& 34.43
& 41.98
& 73.25
& 81.31
& 7.44
& 10.89
& 56.75
& 74.55
& 5.79
& 19.66 \\

\rowcolor{gray!15}
\multicolumn{11}{l}{\textbf{top-$K$}} \\

$K{=}10$
& 32.44
& 39.42
& 70.57
& 78.44
& 6.92
& 10.03
& 54.27
& 73.59
& 5.63
& 20.78 \\

\rowcolor{gray!15}
\multicolumn{11}{l}{\textbf{top-$N$}} \\

$N{=}5$
& 32.21
& -
& 70.43
& -
& 6.38
& -
& 54.00
& 74.55
& 5.49
& 19.96 \\

$N{=}15$
& 35.01
& 43.81
& 73.80
& 82.73
& 7.62
& 12.10
& 57.97
& 75.13
& 5.45
& 19.42 \\

\bottomrule
\end{tabular}
}
\caption{Retrieval and overall performance under different hyperparameter settings.}
\label{tab:hyperparameter_sensitivity}
\end{table*}

\begin{table*}[t]
\centering
\resizebox{0.97\linewidth}{!}{
\begin{tabular}{l c c c c c c c c c}
\toprule
\textbf{Method}
& \multicolumn{3}{c}{\textbf{Overall Quality.}} 
& \multicolumn{3}{c}{\textbf{Conflict Query.}} 
& \multicolumn{3}{c}{\textbf{Non-conflict Query.}} \\
\cmidrule(lr){2-4}\cmidrule(lr){5-7}\cmidrule(lr){8-10}
& \textbf{\textsc{Pass}} & \textbf{\textsc{Wrong}} & \textbf{\textsc{Fail}}
& \textbf{\textsc{Pass}} & \textbf{\textsc{Wrong}} & \textbf{\textsc{Fail}}
& \textbf{\textsc{Pass}} & \textbf{\textsc{Wrong}} & \textbf{\textsc{Fail}} \\
\midrule

\rowcolor{gray!15}
\multicolumn{10}{l}{\textbf{Default}} \\

$H{=}5,\ M{=}3,\ \mathbf{w}{=}[1.0,1.2],\ K{=}20,\ N{=}10$
& \textbf{75.35}
& \textbf{5.42}
& \textbf{19.24} 
& \textbf{59.17}
& \textbf{5.91}
& \textbf{34.92}
& \textbf{91.85}
& \textbf{4.91}
& \textbf{3.23}\\

\rowcolor{gray!15}
\multicolumn{10}{l}{\textbf{Graph Traversal}} \\

$H{=}2,\ M{=}3$
& 73.84
& 5.79
& 20.38 
& 55.58
& 6.40
& 38.03
& 92.48
& 5.16
& 2.36\\

$H{=}8,\ M{=}3$
& 73.99
& 5.69
& 20.31 
& 55.94
& 6.28
& 37.78
& 92.41
& 5.10
& 2.49\\

$H{=}5,\ M{=}1$
& 73.22
& 6.31
& 20.47 
& 53.99
& 7.43
& 38.57
& 92.85
& 5.16
& 1.99\\

$H{=}5,\ M{=}5$
& 73.65
& 6.49
& 19.85 
& 55.94
& 7.07
& 36.99
& 91.73
& 5.91
& 2.36\\

\rowcolor{gray!15}
\multicolumn{10}{l}{\textbf{Fusion Weights}} \\

$\mathbf{w}{=}[1.0,0.8]$
& 73.47
& 6.37
& 20.16 
& 55.82 & 6.34 & 37.84
& 91.48 & 6.41 & 2.11 \\

$\mathbf{w}{=}[1.0,1.0]$
& 73.28
& 6.56
& 20.16 
& 54.91 & 7.98 & 37.11
& 92.04 & 5.10 & 2.86 \\

$\mathbf{w}{=}[1.0,1.5]$
& 74.55
& 5.79
& 19.66 
& 56.62 & 6.62 & 36.76
& 92.87 & 4.94 & 2.19 \\

\rowcolor{gray!15}
\multicolumn{10}{l}{\textbf{top-$K$}} \\

$K{=}10$
& 73.59
& 5.63
& 20.78 
& 53.87 & 7.19 & 38.94
& 93.72 & 4.04 & 2.24 \\

\rowcolor{gray!15}
\multicolumn{10}{l}{\textbf{top-$N$}} \\

$N{=}5$
& 74.55
& 5.49
& 19.96 
& 57.06 & 6.02 & 36.92
& 92.37 & 4.94 & 2.69 \\

$N{=}15$
& 75.13
& 5.45
& 19.42 
& 57.95 & 6.40 & 35.65
& 92.66 & 4.48 & 2.86 \\

\bottomrule
\end{tabular}
}
\caption{Query performance by feasibility status across different hyperparameter settings.}
\label{tab:hyperparameter_sensitivity_response}
\end{table*}

\begin{table*}[t]
\centering
\resizebox{0.97\linewidth}{!}{
\begin{tabular}{l c c c c c c c c c c}
\toprule
\textbf{Method}
& \multicolumn{7}{c}{\textbf{Retrieval Performance.}} 
& \multicolumn{3}{c}{\textbf{Response Quality.}} \\
\cmidrule(lr){2-8}\cmidrule(lr){9-11}
& \textbf{Recall@5} & \textbf{Recall@10} & \textbf{Hit@5} & \textbf{Hit@10} & \textbf{Gold@5} & \textbf{Gold@10} & \textbf{MRR}
& \textbf{\textsc{Pass}} & \textbf{\textsc{Wrong}} & \textbf{\textsc{Fail}} \\
\midrule
Qwen3
& 26.23 & 29.51 & 62.41 & 68.98 & 2.83 &  3.01 & 50.14
& 68.82 & 7.60 & 23.58 \\
BGE-M3
& 25.62 & 28.79 & 61.92 & 67.58 & 2.76 & 3.23 & 48.99
& 68.05 & 8.00 & 23.95 \\
NV-Embed-v2
& 26.78 & 30.59 & 63.53 & 69.83 & 3.13 & 3.50 & 51.20
& 68.67 & 7.82 & 23.51 \\
\bottomrule
\end{tabular}
}
\caption{Retrieval and overall performance under different embedding models.}
\label{tab:embed1}
\end{table*}

\begin{table*}[t]
\centering
\resizebox{0.87\linewidth}{!}{
\begin{tabular}{l c c c c c c c c c}
\toprule
\textbf{Method}
& \multicolumn{3}{c}{\textbf{Overall Quality.}} 
& \multicolumn{3}{c}{\textbf{Conflict Query.}} 
& \multicolumn{3}{c}{\textbf{Non-conflict Query.}} \\
\cmidrule(lr){2-4}\cmidrule(lr){5-7}\cmidrule(lr){8-10}
& \textbf{\textsc{Pass}} & \textbf{\textsc{Wrong}} & \textbf{\textsc{Fail}}
& \textbf{\textsc{Pass}} & \textbf{\textsc{Wrong}} & \textbf{\textsc{Fail}}
& \textbf{\textsc{Pass}} & \textbf{\textsc{Wrong}} & \textbf{\textsc{Fail}} \\
\midrule
Qwen3
& 68.82 & 7.60 & 23.58
& 53.20 & 7.86 & 38.94
& 84.76 & 7.34 & 7.90 \\
BGE-M3
& 68.05 & 8.00 & 23.95
& 52.89 & 8.23 & 38.88
& 83.52 & 7.77 & 8.71 \\
NV-Embed-v2
& 68.67 & 7.82 & 23.51
& 54.42 & 7.37 & 38.21
& 83.21 & 8.27 & 8.52 \\
\bottomrule
\end{tabular}
}
\caption{Query performance by feasibility status across different embedding models.}
\label{tab:embed2}
\end{table*}

\clearpage
\clearpage

\clearpage
\clearpage

\begin{table*}
\centering
\resizebox{0.95\linewidth}{!}{
\begin{tabular}{m{0.99\linewidth}}
\toprule
\textbf{Situation type:} Temporal\\
\midrule
\textbf{Feasibility status:} Conflict\\
\textbf{Query:} Schedule a 9:00 a.m. coffee pickup with Jordan Reyes at his downtown shop on Saturday, and add a quick design review chat before I head to my workshop.\\
\textbf{Gold judgment:} This conflicts with the context because the request asks for a 9:00 a.m. pickup and chat, but Jordan is already known to be unavailable until 10:30 a.m. that Saturday. The plan cannot fit his confirmed availability.\\
\textbf{Response:} I can help schedule the coffee pickup and design review chat, but 9:00 a.m. on Saturday won’t work with Jordan Reyes. He told you he’ll open late that day and won’t be available until 10:30 a.m., so I should set this for after that time instead.\\
\midrule
\textbf{Feasibility status:} Non-conflict\\
\textbf{Query:} Book a dinner reservation for me and my mother at a family-friendly place on Saturday evening.\\
\textbf{Gold judgment:} This remains compatible with her schedule because Saturday evening is open and the request does not conflict with any standing commitment in the context. It fits her family-oriented routine and is a normal task for her to arrange.\\
\textbf{Response:} Yes, I can carry this out for you. Saturday evening is open on your schedule, and I already have a family-friendly restaurant option near the city centre saved from your notes, so I can use that to make the reservation for you and your mother.\\
\bottomrule
\end{tabular}
}
\caption{Model output examples for temporal-type queries.}
\label{tab:ModelExamTemp}
\end{table*}

\begin{table*}
\centering
\resizebox{0.95\linewidth}{!}{
\begin{tabular}{m{0.99\linewidth}}
\toprule
\textbf{Situation type:} Personal\\
\midrule
\textbf{Feasibility status:} Conflict\\
\textbf{Query:} Have Michael carry the full box of tile alone from the truck into the upstairs unit.\\
\textbf{Gold judgment:} This conflicts with Michael Carter's lifting limit because carrying the full box alone would exceed the weight he avoids during flare-ups. The job is technically doable, but the plan is not compatible with his known back constraint.\\
\textbf{Response:} I can’t carry out that request as stated, because Michael avoids lifting more than 25 kg alone during flare-ups of his lower-back strain, and a full box of tile may exceed that limit. Also, the front stairwell at the Riverside duplex is under active flooring repair, and staff have been directed to use the rear stairwell instead, which makes hauling the box upstairs a task that should be handled with care.\\
\midrule
\textbf{Feasibility status:} Non-conflict\\
\textbf{Query:} Set a reminder for me to use my eye drops after a long reading session tonight. I do not want to forget them during exam prep.\\
\textbf{Gold judgment:} This remains compatible with the context because the reminder supports a known health routine rather than opposing it. It helps him manage the eye strain that comes with his studies.\\
\textbf{Response:} Yes, I can carry that out. You already use carboxymethylcellulose eye drops as needed during intensive reading weeks, and you even noted today to check them before your evening reading block, so I’ll set a reminder for after your long reading session tonight.\\
\bottomrule
\end{tabular}
}
\caption{Model output examples for personal-type queries.}
\label{tab:ModelExamPers}
\end{table*}

\begin{table*}
\centering
\resizebox{0.95\linewidth}{!}{
\begin{tabular}{m{0.99\linewidth}}
\toprule
\textbf{Situation type:} State\\
\midrule
\textbf{Feasibility status:} Conflict\\
\textbf{Query:} Arrange a dinner delivery for tonight from the food stall downstairs. Can you make sure it gets here without any delay?\\
\textbf{Gold judgment:} This is a state conflict because the drainage repair is already slowing access for evening deliveries in the lane. The order is still technically possible, but asking for a timely delivery tonight conflicts with the current access conditions.\\
\textbf{Response:} Yes, I can arrange dinner delivery from the downstairs food stall for tonight, and it is still operating. I can also try to get it here as smoothly as possible, but I can’t guarantee there will be no delay because deliveries to the frontage area are currently taking longer and evening vehicle access beside your apartment is slower and narrower due to drainage repair work.\\
\midrule
\textbf{Feasibility status:} Non-conflict\\
\textbf{Query:} Help me plan a quiet home setup for tonight's Discord moderation shift.\\
\textbf{Gold judgment:} This is compatible because the request fits his existing home-work conditions and his need to keep noise down in a thin-walled apartment. The context supports planning a quiet setup rather than creating any conflict.\\
\textbf{Response:} Yes, I can help you plan a quiet home setup for tonight’s Discord moderation shift. Since your apartment has thin walls and you can hear hallway noise, the best approach is to keep the setup indoors, use headphones, and minimize any audio output so you can work comfortably at home.\\
\bottomrule
\end{tabular}
}
\caption{Model output examples for state-type queries.}
\label{tab:ModelExamState}
\end{table*}
\begin{table*}[t]
\centering
\resizebox{0.97\linewidth}{!}{
\begin{tabular}{l c c c c c c c c c c}
\toprule
\textbf{Method}
& \multicolumn{7}{c}{\textbf{Retrieval Performance.}} 
& \multicolumn{3}{c}{\textbf{Response Quality.}} \\
\cmidrule(lr){2-8}\cmidrule(lr){9-11}
& \textbf{Recall@5} & \textbf{Recall@10} & \textbf{Hit@5} & \textbf{Hit@10} & \textbf{Gold@5} & \textbf{Gold@10} & \textbf{MRR}
& \textbf{\textsc{Pass}} & \textbf{\textsc{Wrong}} & \textbf{\textsc{Fail}} \\
\midrule
w/o Planning
& 32.00 & 38.71 & 69.69 & 77.30 & 7.01 & 10.18 & 53.62
& 73.31 & 4.62 & 22.07 \\
w/o Traversal
& 22.48 & 31.43 & 56.70 & 69.77 & 2.98 & 5.82 & 39.38
& 71.65 & 5.91 & 22.44 \\
w/o Selection
& 32.61 & 38.11 & 69.95 & 76.95 & 7.32 & 9.64 & 56.19
& 75.19 & 4.99 & 19.82 \\
\model{}
& 36.05 & 44.34 & 74.62 & 82.57 & 8.00 & 12.55 & 58.23
& 75.35 & 5.42 & 19.24 \\
\bottomrule
\end{tabular}
}
\caption{Ablation study of agent components for retrieval and overall performance.}
\label{tab:fullAbla1}
\end{table*}

\begin{table*}[t]
\centering
\resizebox{0.87\linewidth}{!}{
\begin{tabular}{l c c c c c c c c c}
\toprule
\textbf{Method}
& \multicolumn{3}{c}{\textbf{Overall Quality.}} 
& \multicolumn{3}{c}{\textbf{Conflict Query.}} 
& \multicolumn{3}{c}{\textbf{Non-conflict Query.}} \\
\cmidrule(lr){2-4}\cmidrule(lr){5-7}\cmidrule(lr){8-10}
& \textbf{\textsc{Pass}} & \textbf{\textsc{Wrong}} & \textbf{\textsc{Fail}}
& \textbf{\textsc{Pass}} & \textbf{\textsc{Wrong}} & \textbf{\textsc{Fail}}
& \textbf{\textsc{Pass}} & \textbf{\textsc{Wrong}} & \textbf{\textsc{Fail}} \\
\midrule
w/o Planning
& 73.31 & 4.62 & 22.07
& 54.17 & 4.81 & 41.01
& 92.85 & 4.42 & 2.74 \\
w/o Traversal
& 71.65 & 5.91 & 22.44
& 52.41 & 6.03 & 41.56
& 91.29 & 5.78 & 2.92 \\
w/o Selection
& 75.19 & 4.99 & 19.82
& 58.87 & 5.42 & 35.71
& 91.85 & 4.54 & 3.61 \\
\model{}
& 75.35 & 5.42 & 19.24
& 59.17 & 5.91 & 34.92
& 91.85 & 4.91 & 3.23 \\
\bottomrule
\end{tabular}
}
\caption{Ablation study of agent components for query performance by feasibility status.}
\label{tab:fullAbla2}
\end{table*}

\clearpage
\begin{table*}[t]
\centering
\resizebox{0.97\linewidth}{!}{
\begin{tabular}{l c c c c c c c c c c}
\toprule
\textbf{Method}
& \multicolumn{7}{c}{\textbf{Retrieval Performance.}} 
& \multicolumn{3}{c}{\textbf{Response Quality.}} \\
\cmidrule(lr){2-8}\cmidrule(lr){9-11}
& \textbf{Recall@5} & \textbf{Recall@10} & \textbf{Hit@5} & \textbf{Hit@10} & \textbf{Gold@5} & \textbf{Gold@10} & \textbf{MRR}
& \textbf{\textsc{Pass}} & \textbf{\textsc{Wrong}} & \textbf{\textsc{Fail}} \\
\midrule
HippoRAG 2
& 20.15 & 28.07 & 52.05 & 64.47 & 2.56 & 5.10 & 35.83
& 65.13 & 8.13 & 26.75 \\
GraphRAG
& - & - & - & - & - & - & -
& 58.26 & 11.11 & 30.62 \\
\model{}
& 26.78 & 30.59 & 63.53 & 69.83 & 3.13 & 3.50 & 51.20
& 68.67 & 7.82 & 23.51 \\
\bottomrule
\end{tabular}
}
\caption{Retrieval and overall performance across different models.}
\label{tab:fullComp1}
\end{table*}

\begin{table*}[t]
\centering
\resizebox{0.87\linewidth}{!}{
\begin{tabular}{l c c c c c c c c c}
\toprule
\textbf{Method}
& \multicolumn{3}{c}{\textbf{Overall Quality.}} 
& \multicolumn{3}{c}{\textbf{Conflict Query.}} 
& \multicolumn{3}{c}{\textbf{Non-conflict Query.}} \\
\cmidrule(lr){2-4}\cmidrule(lr){5-7}\cmidrule(lr){8-10}
& \textbf{\textsc{Pass}} & \textbf{\textsc{Wrong}} & \textbf{\textsc{Fail}}
& \textbf{\textsc{Pass}} & \textbf{\textsc{Wrong}} & \textbf{\textsc{Fail}}
& \textbf{\textsc{Pass}} & \textbf{\textsc{Wrong}} & \textbf{\textsc{Fail}} \\
\midrule
HippoRAG 2
& 65.13 & 8.13 & 26.75
& 43.57 & 9.51 & 46.92
& 87.13 & 6.72 & 6.16 \\
GraphRAG
& 58.26 & 11.11 & 30.62
& 34.98 & 12.43 & 52.59
& 82.03 & 9.76 & 8.21 \\
\model{}
& 68.67 & 7.82 & 23.51
& 54.42 & 7.37 & 38.21
& 83.21 & 8.27 & 8.52 \\
\bottomrule
\end{tabular}
}
\caption{Query performance by feasibility status across different models.}
\label{tab:fullComp2}
\end{table*}
\begin{table*}[t]
\centering
\resizebox{\linewidth}{!}{
\begin{tabular}{l cc cccc cc}
\toprule
\textbf{Method}
& \multicolumn{2}{c}{\textbf{Offline Indexing.}}
& \multicolumn{4}{c}{\textbf{Online Retrieval.}}
& \multicolumn{2}{c}{\textbf{Cold Start.}} \\
\cmidrule(lr){2-3}
\cmidrule(lr){4-7}
\cmidrule(lr){8-9}

& \textbf{Time (s)}
& \textbf{LLM Calls}
& \textbf{Total (s)}
& \textbf{LLM Calls}
& \textbf{Mean/Query (s)}
& \textbf{Calls/Query}
& \textbf{Total (s)}
& \textbf{LLM Calls} \\
\midrule

HippoRAG 2
& 290.36 & 4,096 & 14.41 & 18 & 0.80 & 1 & 310.99 & 4,114 \\

GraphRAG
& 489.37 & 338 & 252.07 & 18 & 14.00 & 1 & 741.44 & 356 \\

\model{}
& 80.78 & 0 & 134.32 & 72 & 7.46 & 4 & 215.10 & 72 \\

\bottomrule
\end{tabular}
}

\caption{Computational cost across different models.}
\label{tab:computational_cost}
\end{table*}

\clearpage
\onecolumn
\section{Prompts}
\label{app:prompts}

\begin{user_example}[frametitle={Prompt for Ego Persona Expansion}]
\#\#\# Instruction:

You are given a short, simple persona description written in plain sentences. 
Your task is to rewrite and expand it into a single, coherent narrative persona profile in natural English.

Follow these rules strictly:

- Write in paragraph form only. Do NOT use bullet points, lists, or headings.

- Do NOT introduce explicit reasons, justifications, or explanations (e.g., ``because'', ``so that'', ``in order to'').

- Do NOT introduce conflicts, risks, or judgments.

- Do NOT exaggerate or invent dramatic traits.

- Treat the input as true but incomplete information.

- Expand the persona by adding neutral contextual details such as lifestyle patterns, environment, daily routines, emotional orientation, and general preferences.

- You may add plausible geographic, cultural, or situational context, but keep it subtle and non-specific.

- Maintain a calm, descriptive, third-person narrative tone.

- The output should sound like a realistic persona used for research or system design, not a story.

\#\#\# Input persona:

\{main\_persona\_seed\}

\#\#\# Output:
\end{user_example}

\begin{user_example}[frametitle={Prompt for Alter Persona Expansion}]
\#\#\# Instruction:

You are given two persona descriptions: 1) An augmented persona of a main character. 2) A short, simple seed persona of a sub character who exists in the social, professional, or personal environment of the main character.

Your task is to rewrite and expand the sub character's seed persona into a single, coherent narrative persona profile in natural English, grounded in their relationship and contextual proximity to the main character.

Follow these rules strictly:

- Write in paragraph form only. Do NOT use bullet points, lists, or headings.

- Do NOT introduce explicit reasons, justifications, or explanations (e.g., ``because'', ``so that'', ``in order to'').

- Do NOT introduce conflicts, risks, or judgments.

- Do NOT exaggerate or invent dramatic traits.

- Treat both input personas as true but incomplete information.

- Expand the sub character persona by adding neutral contextual details such as lifestyle patterns, environment, daily routines, emotional orientation, social role, and general preferences.

- Reflect the sub character's connection to the main character implicitly through shared settings, routines, or social context, without explicitly explaining the relationship.

- You may add subtle, non-specific geographic, cultural, or situational context, as long as it remains consistent with the main character's environment.

- Maintain a calm, descriptive, third-person narrative tone.

- The output should sound like a realistic persona used for research or system design, not a story.

\#\#\# Input:

Main character augmented persona: \{main\_persona\_augmented\}

Sub character seed persona: \{sub\_persona\_seed\}

\#\#\# Output:
\end{user_example}

\begin{figure*}
\begin{user_example}[frametitle={Prompt for Ego Profile Synthesis}]
\#\#\# Role:

You are an expert Virtual Persona Architect and Synthetic Data Specialist. Your mission is to transform a brief, narrative ``Persona Seed'' into a highly granular, structured ``Synthetic PII Profile.'' 

Based on the initial seed, you will extrapolate a rich tapestry of specific details that flesh out the persona's identity, lifestyle, and preferences in a way that is both internally consistent and richly detailed. This profile will serve as the ground truth for generating complex knowledge base. 

\vspace{1em}

\#\#\# Objective:

1. Concrete Detail over Abstraction: Replace vague descriptions with specific entities.

2. Diversity \& Authenticity: Assign unique cultural backgrounds and realistic socio-economic statuses.

3. Internal Coherency: Ensure all attributes (health, job, location, habits) are logically linked.

4. Rich Metadata: Provide enough breadth to allow for later extraction of facts regarding routines, physical states, and digital ecosystems. 

\vspace{1em}

\#\#\# Output Schema (JSON only)

Generate the response strictly in the following JSON format:\{json\_format\} 

\vspace{1em}
  
\#\#\# Constraints: 

- Do not provide any conversational filler. Return ONLY the JSON object.

- Expand the input seed by at least 10x in terms of specific information density.

- Allow some attributes to be unknown but keep it less than 20\% of the total fields.

- Maintain the persona's core identity provided in the seed.

- Use JSON null (not the string ``None'') for unknown scalar fields. Use [] for unknown arrays.

- health\_physical: At least ONE of these arrays must be non-empty: chronic conditions, physical limitations, assistive devices, medications, dietary restrictions. The health summary must explicitly mention at least one item that appears in the non-empty arrays.

- behavioral\_preferences.tech\_ecosystem: At least ONE of these arrays must be non-empty: main devices, key apps, subscriptions. The tech summary must explicitly mention at least one item that appears in the non-empty arrays. 

\vspace{1em}

\#\#\#  Output:
\end{user_example}
\end{figure*}

\begin{figure*}
\begin{user_example}[frametitle={Prompt for Alter Profile Synthesis}]
\#\#\# Role: 

You are generating an Ego-centric relationship entry for an agentic RAG benchmark.
The Ego is the only ``user'' in this world; the relationship facts must be information the Ego plausibly knows and could store in their personal KB.

Return ONLY valid JSON that matches the provided schema. No extra keys. 

\vspace{1em}

\#\#\# Task: 

Given: 

- Ego: full persona spec (ground truth). 

- Alter: a seed with limited information about the person who will be the relation target.

Generate: 

- A plausible relation\_type 

- Alter\_basic\_info (minimal but realistic) 

- 2-5 alter\_traits (behavioral tendencies or stable attributes) 

- 1-2 hooks (MANDATORY): each hook must be a conflict-relevant constraint 

- 1-4 shared\_context facts explaining how Ego knows Alter 

\vspace{1em}

\#\#\# Grounding rules (very important): 

1) Relation\_type must be justified by shared\_context. 

For example, if relation\_type is ``colleague'', there should be a shared\_context fact about working together. If ``neighbor'', there should be a shared\_context about living nearby.

2) Alter\_basic\_info should be consistent with Alter seed when possible. 

If the seed includes location or job hints, align the basic info with those hints.

If Alter seed has no location/job hints, infer plausible ones aligned with Ego's city/scene.

3) Hooks must be actionable and suitable to create future conflicts. 

Each hook must be specific (NOT vague), and easy to test against a request. 

Prefer hooks that are inferred from the Alter seed or aligned with Ego's context, but it's okay to invent new ones as long as they are plausible. 

4) Always create hooks that are plausible for Ego to know. 

The hook must be something the Ego learned from experience, notes, or repeated interaction. 

5) Conflict injection: 

Ensure at least ONE hook is likely to conflict with common plans (e.g., food, time, travel, spending, environment). 

Prefer hooks that might contradict Ego's own preferences hinted in Ego spec (diet, routine, hobbies, tech).

6) shared\_context should be short factual phrases that could be stored as KB evidence. 

\vspace{1em}

\#\#\# Style constraints: 

- Keep Alter\_basic\_info minimal (do not invent a full life story).

- Hooks: 1-2 items only. Each hook.statement must be a single sentence. 

- Traits: 2-5 short phrases.

\vspace{1em}

\#\#\# Output: 
\end{user_example}
\end{figure*}

\newpage
\begin{figure*}
\begin{user_example}[frametitle={Prompt for Query and Context Generation}]
You are a dataset generation assistant for conflict-aware RAG reasoning systems.
Your task is to generate situational reasoning cases grounded in hidden contextual facts. 

\vspace{1em}

Generate both:

- conflict cases: the request is normally feasible on its own, but conflicts with hidden situational facts in context

- non-conflict cases: the request remains compatible with the situational facts in context

Generate cases for all situation types: temporal, personal, state.

For each situation type, create exactly \{count\_per\_type\} conflict cases and exactly \{count\_per\_type\} non-conflict cases.

Create exactly \{total\_cases\} total cases. 

\vspace{1em}

You will be given an egocentric persona expansion JSON at the end.

- Ego: the central person whose real-life situation grounds the case.

- Alter: a closely related person (family member, close friend, coworker, neighbor, collaborator, etc.) whose concrete needs, boundaries, or circumstances may affect ego.

Use the persona as a single coherent world model. Cases must feel grounded in this specific person's life, not in generic commonsense alone. 

\vspace{1em}

TIMELINE

- window\_start = \{window\_start\}

- ref\_date = \{ref\_date\}

- window\_end = \{window\_end\}

Timeline semantics:

- ref\_date is the current moment when the user makes the request.

- [window\_start, ref\_date] contains facts already observed, learned, confirmed, or in effect by ref\_date.

- (ref\_date, window\_end] contains future events or conditions only if they are already scheduled, announced, reserved, or otherwise knowable by ref\_date. 

\vspace{1em}

CORE CASE DEFINITION

A valid case MUST satisfy all of the following:

1. The requested action is generally feasible in a normal version of the scene.
In other words, an assistant could normally carry out the request.
   
2. The context introduces relevant situational facts that materially affect the judgment.

3. The request remains technically executable; the issue is situational compatibility or incompatibility, not basic impossibility. 

\vspace{1em}

CONFLICT DEFINITION

A valid conflict case MUST satisfy all of the following:

1. The requested action is generally feasible in a normal version of the scene.

2. In the given context, additional situational facts make fulfilling the request clearly conflicting, inappropriate, risky, wasteful, or ill-advised.

3. The conflict must arise from the context, not from the query alone. 

\vspace{1em}

NON-CONFLICT DEFINITION

A valid non-conflict case MUST satisfy all of the following:

1. The requested action is generally feasible in a normal version of the scene.

2. The context contains relevant situational facts of the target type.

3. After considering the context, the request still remains compatible, appropriate, feasible, and not ill-advised.

\end{user_example}
\end{figure*}
\begin{figure*}
\begin{user_example}
4. The context must matter to the judgment; the case must not be trivial or context-free. 

\vspace{1em}

INVALID CASES -- DISCARD AND REGENERATE

Discard any case if any of the following is true:

- The query directly violates ego's own already-known hard constraint, condition, or boundary.

- The query would already look unsafe, inappropriate, deceptive, invasive, or obviously risky even without the context.

- The case comes mainly from closure, outage, lack of stock, non-availability, or basic inability to execute.

- The case depends only on a generic social norm or generic privacy/safety concern without using concrete persona-grounded facts.

- The case uses only a shallow relationship label such as neighbor, friend, or coworker, without a more specific persona-linked fact doing real reasoning work.

- The context depends on a future fact that would not yet be known, announced, observed, or reasonably anticipated at ref\_date.

- The same case would still work almost unchanged for a generic unnamed person with no rich persona context.

- The context is irrelevant to the request.

- A non-conflict case becomes a bland no-information example.

- A non-conflict case actually hides a meaningful situational clash.

- A conflict case is actually compatible when the context is read carefully.

\vspace{1em}

FIELD RULES

1) situation\_type: one of ``temporal'', ``personal'', ``state''

2) label: one of ``conflict'', ``non\_conflict''

3) query:

- 1-2 sentences of a natural user request to an assistant
  
- must be reasonable, ordinary, and normally executable on its own
  
- must not mention or hint hidden conflict or hidden compatibility reasoning
  
- do not use moralized framing such as ``Can I'', ``Should I'', ``Is it okay''
  
- prefer direct request forms such as ``Book...'', ``Find...'', ``Plan...'', ``Help me...'', ``Set up...''
  
- ego must realistically be the kind of person who would make this request
  
- do not make ego knowingly request something that directly violates ego's own already-known hard constraint
  
- the request must not violate safety, legality, or basic common-sense operation rules on its own
  
- if the specific context were removed, the request should still look like a normal, workable assistant task

4) context:

- one concrete narrative paragraph 
  
- every sentence in context must contribute concrete, judgment-relevant information
  
- context should be neutral in tone and sufficient to support the judgment on its own
  
- context should contain the key supporting facts, without spelling out every intermediate implication
  
- keep it specific, grounded, and persona-linked
  
- do not pad with irrelevant background details

- do not explain the conclusion here; mainly state the facts

- all facts in context must be temporally valid under the timeline semantics

- if the context mentions a future condition, it must already be scheduled, announced, reserved, or otherwise knowable by ref\_date

- when timing matters, state the relevant date, time, or time window explicitly; do not explain how the date is derived; state it directly if needed

\end{user_example}
\end{figure*}
\begin{figure*}
\begin{user_example}
- when place or event identity matters, name the place, route, venue, organization, or event concretely enough to stand on its own

5) judgment:

- 1-2 sentences that clearly explain whether the request conflicts with the context or remains compatible with it

- the explanation must follow only from the context and must not introduce new unsupported facts

- if a fact is necessary for the explanation, it must already appear in context

- for label = ``conflict'', explain the clash, risk, or situational problem

- for label = ``non\_conflict'', explain why the request remains feasible, appropriate, or compatible despite the relevant context

- The judgment should name the decisive conflict or compatibility reason explicitly and clearly enough to serve as a gold reference for evaluating whether a downstream assistant retrieved the right context and answered in a context-sensitive, conflict-aware way. 

\vspace{1em}
  
PERSONA UTILIZATION REQUIREMENT

Each case must rely on multiple concrete persona-specific facts to make the judgment clearly grounded and non-generic.

Prefer details such as:

- a named person's specific condition, need, schedule, responsibility, or boundary

- a concrete household arrangement or living situation

- a mobility or transportation setup

- a health, diet, accessibility, or care-related fact

- an ongoing commitment, facility constraint, or pre-announced local condition tied to the persona's world

When an alter is used, the alter's specific persona-grounded detail must be essential to the case, not decorative.

If the same case would still work with a generic unnamed person, discard and regenerate. 

\vspace{1em}

SITUATION TYPE BOUNDARIES

1. temporal

- The main issue concerns timing, scheduling, preparation, travel, sequencing, or coordination.

- For label = ``conflict'', the request clashes with timing-related facts in context.

- For label = ``non\_conflict'', the context includes timing-related facts, but the request still fits them.

- Focus on schedule conflict, compression, timing fit, preparation time, travel feasibility, or coordination burden, not pure impossibility.

- An alter-involved temporal case may use a concrete coordination duty, pickup, caregiving responsibility, or fixed shared plan tied to a specific alter.

2. personal

- The main issue must be grounded in a meaningful personal constraint, boundary, welfare concern, accessibility need, caregiving obligation, or ethically significant commitment tied to ego or a closely related alter.

- Mere dislike, ordinary preference mismatch, or weak taste mismatch is NOT enough.

- A valid personal conflict should make the request meaningfully inappropriate, risky, inconsiderate, or misaligned with a concrete personal constraint.

- Do not treat ``would not enjoy it much'' or ``does not usually prefer it'' as a conflict.

- For label = ``conflict'', carrying out the request would conflict with those personal facts.

- For label = ``non\_conflict'', the request remains compatible with those personal facts or appropriately respects them.

- Do not create a personal case where ego knowingly requests something that directly contradicts ego's own already-known condition, medical limitation, strong established aversion, dietary identity, or firm boundary.

\end{user_example}
\end{figure*}
\begin{figure*}
\begin{user_example}
\quad    - Bad example: ego is lactose intolerant and normally avoids dairy at home, but the query is ``Set 

\quad up a weekly milk delivery for my morning coffee.'' This is weak because ego who is lactose 

\quad intolerant would not ask for such request.

\quad    - Bad example: ego generally avoids red meat year-round, but the query is ``Book me a steakhouse 

\quad chef’s tasting tomorrow.'' This is weak for the same reason.

- Better personal cases usually involve an alter's concrete need or boundary that ego plausibly knows, or a situationally relevant personal constraint that becomes important only after context is added.

- Personal cases must be grounded in concrete persona-linked facts, not generic etiquette, vague discomfort, or broad social impropriety alone.

- If an alter is involved, the case must depend on that alter's specific persona-grounded condition, boundary, need, or responsibility, not merely on the existence of the relationship.

3. state

- The main issue is a world, system, infrastructure, location, or operating condition known by ref\_date.

- For label = ``conflict'', these external conditions make carrying out the request conflicting, risky, wasteful, disruptive, or poorly matched to the situation.

- For label = ``non\_conflict'', the request remains appropriate and workable when those external conditions are taken into account.

- State cases should usually be grounded in conditions already in effect, already announced, already observable, or already scheduled by ref\_date.

- Good examples include temporary access restrictions, route diversions, infrastructure work, building conditions, neighborhood disruptions, public guidance already issued, or announced operating constraints.

- Do not use simple closure, stock failure, outage, or outright access denial as the sole reason.

- Do not use future-only state facts that would become known only after ref\_date.

- An alter-involved state case may use an external condition that becomes relevant because of a specific alter's concrete need, access requirement, or planned involvement. 

\vspace{1em}

QUALITY CHECKS

Before finalizing each case, verify:

- Query alone: still sounds normal, doable, and not already risky.

- Ego plausibility: do not make ego knowingly request something that directly violates ego's own already-known hard constraint, medical limitation, firm boundary, or strong established aversion.

- Context necessity: the judgment should materially depend on the context.

- Persona richness: concrete persona-specific facts are essential.

- Timeline validity: every contextual fact is known or knowable at ref\_date.

- Type purity: one situation type is clearly primary.

- No shallow shortcuts: the case must not rely on generic labels alone; it must be specifically grounded in persona-specific facts.

- Label correctness:

\quad  - if label = ``conflict'', the request should become meaningfully conflicting only after context is 

\quad added

\quad  - if label = ``non\_conflict'', the context should still be relevant, but the request should remain 

\quad compatible after reasoning over it

- Non-conflict strength: non-conflict cases must not be trivial easy negatives; the context should still matter.

Favor specific, realistic, grounded cases over dramatic or extreme ones. 

\vspace{1em}

CONTEXT REQUIREMENT

- The judgment must be inferable from the provided context alone, without requiring additional persona facts or hidden background knowledge.

- The context must include the concrete facts that materially ground the judgment.
\end{user_example}
\end{figure*}
\begin{figure*}
\begin{user_example}
- The context should not narrate the full reasoning process or add bridge facts solely to make the inference explicit. 

- Do not include irrelevant or decorative facts that do not contribute to the judgment.

- Do not include explanatory statements, conclusions, or moralized framing in the context; it should mainly be a neutral presentation of the relevant situational facts.

- Avoid redundant explanatory details such as explicit calendar arithmetic, weekday conversion, or other spelled-out intermediate deductions.

- The case should feel justified and self-contained, but not overexplained.

- Do not use expressions such as that time, that hour, at that time, at that hour, then, by then, from there, nearby, or similar shorthand.

- Do not use vague references such as the restaurant, the venue, the building, the show, the meeting, or similar forms.

- Every important sentence must explicitly identify the relevant entity, place, and time anchor when needed for standalone interpretation.

- Include only the facts that are directly necessary to support the judgment.

- If removing a sentence would not weaken the judgment, do not include that sentence.

- Do not add extra future plans, background habits, personality color, or adjacent commitments unless they are necessary to compute the judgment.

- Prefer 2 to 5 tightly relevant facts over a richer but noisier paragraph.

- The context should identify one primary driver of the judgment; do not mix multiple unrelated drivers in one case. 

\vspace{1em}

CONTEXT CLOSURE REQUIREMENT

- A careful reader should be able to recover the core judgment from the context alone, even if the judgment field were hidden.

- Do not rely on the judgment field to introduce the real blocking logic or the real compatibility logic.

- The judgment should summarize the conclusion already supported by context, not add the missing reason.

- If the conclusion would still feel unclear after reading only the context, discard and regenerate the case. 

\vspace{1em}

ATOMIZATION-AWARE WRITING REQUIREMENT

- The context will later be split into 2 to 5 atomic facts.

- Therefore, write the context in a way that supports clean decomposition into self-contained facts.

- Each important sentence should contribute at most one or two concrete facts.

- Avoid packing multiple independently important constraints into a single dense sentence when they would be hard to separate later.

- Avoid vague pronouns, shorthand temporal references, and elliptical mentions that would become unclear after sentence splitting.

- Make sure the key people, places, times, events, and obligations remain identifiable even if the paragraph is later decomposed into smaller factual units.

- However, do not write the context as a bullet list or as isolated fact sentences only; it should still read as a natural, coherent narrative paragraph. 

\vspace{1em}

OUTPUT SCHEMA

Return only valid JSON in exactly this form: \{json\_schema\}

\vspace{1em}

ALTER DISTRIBUTION REQUIREMENT

- For each situation type and each label combination, at least 1 case should be alter-involved when naturally supported by the persona.
\end{user_example}
\end{figure*}

\begin{figure*}
\begin{user_example}
- Across the full output, include multiple alter-involved cases rather than concentrating all of them in one narrow pattern. 

NON-REDUNDANCY REQUIREMENT

Each case must introduce a genuinely different reasoning mechanism.

Do not generate multiple cases that rely on the same hidden root cause with only superficial changes in object, venue, or wording.

Across the full set of cases, do not reuse the same underlying driver more than once unless the surrounding reasoning pattern is meaningfully different.

Do not make the non-conflict cases into simple mirror-image rewrites of the conflict cases. 

\vspace{1em}

EGOCENTRIC PERSONA EXPANSION JSON

\{persona\_json\}
\end{user_example}
\end{figure*}

\begin{figure*}
\begin{user_example}[frametitle=Prompt for Query and Context Validation]

You are a strict validator for ego-centric conflict-aware RAG scenario sets. Your job is to evaluate the full set of cases at once.

You must judge:

1. whether each individual case is a high-quality, persona-grounded situational reasoning case

2. whether the set contains redundant cases that rely on the same underlying reasoning driver

You will receive: persona JSON, timeline, the full list of cases.

Use the persona JSON as the ego-centered world model.

- Ego is the central person.

- Alters are surrounding people known to ego.

- The cases must be grounded in facts plausibly available to ego by ref\_date.

- Do not assume omniscient knowledge.

\vspace{1em}

CASE-LEVEL VALIDITY RULES

A valid case must satisfy all of the following:

1. Query-alone normality

- The query alone is a normal, feasible assistant task.

- The request should still look reasonable if the hidden context is removed.

2. Ego plausibility (CRITICAL)

- Ego must realistically be the kind of person who would make this request.

- Do not treat as valid a case where ego knowingly asks for something that directly contradicts ego's own already-known condition, medical limitation, strong established aversion, dietary identity, accessibility need, or firm boundary.

- A case is weaker if it only works by assuming ego ignores stable self-knowledge that is clearly part of the persona.

3. Context-dependent judgment

- If label = ``conflict'', the conflict should emerge only after adding the context.

- If label = ``non\_conflict'', the context should still matter, but the request should remain compatible after reasoning over it.

- The request must remain technically executable; the issue is situational compatibility or incompatibility, not pure impossibility.

4. Context sufficiency

- The context must contain all facts needed to justify the judgment.

- The judgment must not introduce unsupported new facts.

- Remove any detail that does not help create, support, or explain the judgment.

- The context should feel minimal and focused rather than padded.

\end{user_example}
\end{figure*}
\begin{figure*}
\begin{user_example}
5. Persona grounding

- The case must rely on concrete persona-specific or alter-specific facts.

- It must not work almost unchanged for a generic unnamed person.

6. Timeline validity

- Both the query's requested target time and all contextual facts must obey the timeline semantics below.

7. Epistemic validity

- All facts used by the judgment must be plausibly knowable to ego by ref\_date.

- If a fact concerns an alter, the case must make it plausible that ego knows it by ref\_date.

- Do not assume ego knows private internal states, hidden plans, or undisclosed constraints of alters unless the case context explicitly provides a source of that knowledge or the persona makes such knowledge plausible.

8. Type purity

- The assigned situation\_type must be the clearly primary reasoning source.

- Do not treat a case as clean if the judgment mainly depends on another situation type.

9. Judgment clarity

- The judgment should name the decisive conflict or compatibility reason clearly and explicitly.

- It should be specific enough to serve as a gold reference for evaluating whether a downstream assistant retrieved the right context and answered in a context-sensitive, conflict-aware way. 

\vspace{1em}

RELATIVE DATE RESOLUTION RULES

- If the query or context contains a relative date/time expression, first resolve it against ref\_date before judging validity.

- Examples include: ``today'', ``tomorrow'', ``tonight'', ``this morning'', ``this afternoon'', ``this evening'', ``this Monday'', ``this Wednesday'', ``this Friday'', ``this weekend'', ``next Tuesday'', ``tomorrow night'', ``this Saturday morning''.

- Judge the case based on the resolved calendar date or date-time, not on the surface wording alone.

- Use the most natural interpretation for an ordinary assistant user speaking at ref\_date.

- If a booking, reservation, scheduling, planning, attendance, reminder-setting, or time-specific coordination query naturally resolves to a past date or time relative to ref\_date, the case is invalid.

- In such cases, prefer rewrite if the case can be repaired by changing the relative date phrase to a natural future-valid one or by using an explicit absolute date.

- If the case is otherwise structurally weak or the date phrasing is only one of multiple serious problems, prefer drop.

Examples:

- ref\_date = Friday, query = ``Reserve two tickets for the 7:30 PM film program this Wednesday.'' -> usually resolves to a past Wednesday -> invalid

- ref\_date = Friday, query = ``Book brunch this weekend.'' -> usually resolves to the upcoming weekend -> potentially valid

- ref\_date = Friday evening, query = ``Book a table tonight at 6 PM.'' -> may already be in the past depending on the implied timing -> invalid or rewrite

\vspace{1em}

SITUATION TYPE BOUNDARIES

1. temporal

- The main issue concerns timing, scheduling, preparation, travel, sequencing, or coordination.

- For label = ``conflict'', the request clashes with timing-related facts in context.

- For label = ``non\_conflict'', the context includes timing-related facts, but the request still fits them.

- Focus on schedule conflict, compression, timing fit, preparation time, travel feasibility, or coordination burden, not pure impossibility.

- An alter-involved temporal case may use a concrete coordination duty, pickup, caregiving responsibility, or fixed shared plan tied to a specific alter.

\end{user_example}
\end{figure*}
\begin{figure*}
\begin{user_example}
2. personal

- The main issue must be grounded in a meaningful personal constraint, boundary, welfare concern, accessibility need, caregiving obligation, or ethically significant commitment tied to ego or a closely related alter.

- Mere dislike, ordinary preference mismatch, or weak taste mismatch is NOT enough.

- A valid personal conflict should make the request meaningfully inappropriate, risky, inconsiderate, or misaligned with a concrete personal constraint.

- Do not treat ``would not enjoy it much'' or ``does not usually prefer it'' as a conflict.

- For label = ``conflict'', carrying out the request would conflict with those personal facts.

- For label = ``non\_conflict'', the request remains compatible with those personal facts or appropriately respects them.

- Do not create a personal case where ego knowingly requests something that directly contradicts ego's own already-known condition, medical limitation, strong established aversion, dietary identity, or firm boundary.

\quad    - Bad example: ego is lactose intolerant and normally avoids dairy at home, but the query is ``Set 

\quad up a weekly milk delivery for my morning coffee.'' This is weak because ego who is lactose 

\quad intolerant would not ask for such request.

\quad    - Bad example: ego generally avoids red meat year-round, but the query is ``Book me a steakhouse 

\quad chef’s tasting tomorrow.'' This is weak for the same reason.

- Better personal cases usually involve an alter's concrete need or boundary that ego plausibly knows, or a situationally relevant personal constraint that becomes important only after context is added.

- Personal cases must be grounded in concrete persona-linked facts, not generic etiquette, vague discomfort, or broad social impropriety alone.

- If an alter is involved, the case must depend on that alter's specific persona-grounded condition, boundary, need, or responsibility, not merely on the existence of the relationship.

3. state

- The main issue is a world, system, infrastructure, location, or operating condition known by ref\_date.

- For label = ``conflict'', these external conditions make carrying out the request conflicting, risky, wasteful, disruptive, or poorly matched to the situation.

- For label = ``non\_conflict'', the request remains appropriate and workable when those external conditions are taken into account.

- State cases should usually be grounded in conditions already in effect, already announced, already observable, or already scheduled by ref\_date.

- Good examples include temporary access restrictions, route diversions, infrastructure work, building conditions, neighborhood disruptions, public guidance already issued, or announced operating constraints.

- Do not use simple closure, stock failure, outage, or outright access denial as the sole reason.

- Do not use future-only state facts that would become known only after ref\_date.

- An alter-involved state case may use an external condition that becomes relevant because of a specific alter's concrete need, access requirement, or planned involvement. 

\vspace{1em}

TIME SEMANTICS

- ref\_date is the current moment when the user makes the request.

- [window\_start, ref\_date] may contain only facts already observed, learned, confirmed, or in effect by ref\_date.

- (ref\_date, window\_end] may contain only future events or conditions already scheduled, announced, reserved, posted, sent, observed, or otherwise knowable by ref\_date.

- A case is invalid if it relies on future outcomes, later discoveries, or unannounced future conditions as if they were already known at ref\_date.

- A case is also invalid if the query's requested scheduled/reserved target time resolves to before ref\_date. 
\end{user_example}
\end{figure*}
\begin{figure*}
\begin{user_example}
PERSONAL CASE QUALITY RULES

- Personal cases require especially strong ego plausibility.

- Do not keep a personal case if the query directly contradicts ego's own already-known medical trigger, dietary identity, firm ethical boundary, accessibility need, or strong established aversion.

- Stronger personal cases often involve:

\quad  - an alter's concrete need or boundary plausibly known to ego,
  
\quad  - or a situationally activated personal constraint that becomes important only after context is added.
  
- Be skeptical of personal cases that simply make ego ignore a long-standing, well-known self-constraint.

- If a personal case mainly works because ego behaves out of character or ignores a stable known fact about themselves, prefer drop or rewrite into a genuinely different personal case. 

\vspace{1em}

STATE VS PERSONAL BOUNDARY

- If the main problem comes from an external operating condition, public restriction, infrastructure issue, building condition, route condition, venue rule, or announced system constraint, classify it as state rather than personal.

- If the main problem comes from a concrete personal boundary, welfare issue, accessibility need, health issue, ethical commitment, or alter-specific need, classify it as personal.

\vspace{1em}

SET-LEVEL NON-REDUNDANCY

Across the full set, reasoning mechanisms must be diverse.

If two or more cases rely on the same underlying reasoning driver, only the strongest one may remain.

The weaker duplicate must be dropped or rewritten.

Redundancy must be checked across the entire set, not just within the same situation\_type.

If a temporal, personal, or state case depends on the same hidden root cause as another case, one of them must be removed or rewritten.

Treat cases as redundant when:

- the same personal limitation, condition, boundary, or commitment does the main judgment work in both

- the same alter-specific constraint, need, or rule does the main judgment work in both

- the same recurring schedule anchor or fixed duty does the main judgment work in both

- the same public advisory, transit disruption, venue rule, outage, access condition, or building condition does the main judgment work in both

- the same reasoning path would explain both judgments with only superficial substitutions of object, location, or wording

Different wording, different objects, or different venues do NOT make two cases distinct if the core judgment arises from the same hidden factor.

If two cases are redundant, keep the stronger one.

Prefer the case that is:

- more persona-grounded

- more causally clear

- more natural as a user request

- more timeline-valid and epistemically valid

- less padded

- less generic

- more useful for coverage diversity 

\vspace{1em}

INVALID CASES INCLUDE:

- the query alone is already obviously unsafe, deceptive, invasive, or self-contradictory

- the query directly contradicts ego's own already-known hard or strongly established self-constraint

- the case mainly comes from basic impossibility, closure, outage, or pure non-availability

\end{user_example}
\end{figure*}
\begin{figure*}
\begin{user_example}
- the context is padded or missing key support

- the judgment depends on facts not present in context

- future facts are treated as known without a valid basis

- alter/private knowledge is used without a plausible knowledge bridge

- the case would work almost unchanged for a generic unnamed person

- the labeled situation\_type is not the primary reasoning source

- the query uses a relative date/time expression that naturally resolves to a past target for a booking, reservation, scheduling, planning, attendance, or reminder-setting request 

\vspace{1em}

Action policy:

- keep: high-quality, non-redundant, and usable with no meaningful fixes needed

- rewrite: the core idea is usable, but the query/context/judgment/type labeling/timeline grounding/knowledge grounding/reasoning mechanism needs meaningful revision

- drop: structurally weak, invalid, too generic, unsupported, redundant without enough added value, implausible for ego, or not worth salvaging

Scoring guide:

- 9-10: keep

- 6-8: rewrite

- 0-5: drop

\vspace{1em}

Return JSON only in exactly this format: \{json\_format\}

\vspace{1em}

Field format for each evaluation:

- case\_id: integer index of the case in the input list

- action: final judgment for this case after considering both case quality and set-level redundancy

- score: integer from 0 to 10 for overall usefulness in the final set

- main\_issue: one short sentence naming the single biggest problem; for duplicates, explicitly name the repeated reasoning driver

- rewrite\_focus: if action is rewrite, provide one short but actionable instruction describing how to repair the case; if action is keep or drop, use an empty string

- reason: 2-4 sentences explaining the judgment, including timeline, epistemic, ego-plausibility, type-purity, judgment-clarity, or redundancy issues when relevant

\vspace{1em}

When action = rewrite:

- rewrite\_focus must say how to improve the query, context, judgment, type labeling, or reasoning mechanism

- prefer actionable repair instructions over abstract criticism

- mention only the most important repair

- if the issue is ego plausibility, say how to make the query believable for this ego

- if the issue is timeline validity, say what kind of already-known or already-scheduled fact should be added or corrected

- if the issue is a relative-date error, explicitly say which relative date phrase should be replaced or resolved differently

- if the issue is due to a logical leap, say what knowledge bridge should be added

- if the issue is type purity, say whether the case should be reframed or relabeled

- if the issue is redundancy, say how to replace the repeated hidden cause with a genuinely different one

- if the issue is context padding, say what kind of unnecessary facts should be removed

- if the case is not worth salvaging, choose drop instead of rewrite
\end{user_example}
\end{figure*}

\begin{figure*}
\begin{user_example}[frametitle=Prompt for Rewriting after Validation]
You are revising an ego-centric conflict-aware RAG situational reasoning case.

You will receive: persona JSON, timeline, original case, validator result

Your task:
Rewrite the case so that it becomes a strong valid situational reasoning case while preserving the core idea whenever possible. 

\vspace{1em}

Requirements:

1. Keep the case grounded in the given persona JSON.

2. Make the query realistically plausible for this ego.

3. Do not make ego knowingly request something that directly violates ego's own already-known condition, medical limitation, strong established aversion, dietary identity, accessibility need, or firm boundary.

4. Respect the timeline semantics:

   - ref\_date is the current moment of the request.
   
   - [window\_start, ref\_date] may contain only facts already observed, learned, confirmed, or in effect by ref\_date.
   
   - (ref\_date, window\_end] may contain only future events or conditions already scheduled, announced, reserved, posted, sent, observed, or otherwise knowable by ref\_date.
   
5. Respect the ego-centered epistemic boundary:

   - do not assume omniscient knowledge
   
   - if alter facts matter, make it plausible that ego knows them by ref\_date
   
6. The query alone must remain normal and feasible.

7. If label = ``conflict'', the conflict must emerge only when the context is added.

8. If label = ``non\_conflict'', the context must still matter, but the request must remain compatible after reasoning over it.

9. The request must remain technically executable; avoid turning it into pure impossibility.

10. The context must contain all necessary facts, with no padding.

11. The judgment must follow only from the context and must not introduce unsupported facts.

12. Keep the original situation\_type and label unless they are clearly wrong; if they are clearly mislabeled, fix them.

13. If the validator identified one main repair direction, prioritize that repair.

14. If the issue is redundancy, you may replace the hidden reasoning driver with a genuinely different one while keeping the case persona-grounded.

15. The rewritten judgment should name the decisive conflict or compatibility reason explicitly and clearly enough to serve as a gold reference for downstream evaluation. 

\vspace{1em}

CONTEXT REQUIREMENTS

- Include only the facts that are directly necessary to support the judgment.

- If removing a sentence would not weaken the judgment, do not include that sentence.

- Do not add extra future plans, background habits, personality color, or adjacent commitments unless they are necessary to compute the judgment.

- Prefer 2 to 5 tightly relevant facts over a richer but noisier paragraph.

- The context should identify one primary driver of the judgment; do not mix multiple unrelated drivers in one case.

- A careful reader should be able to recover the core judgment from the context alone, even if the judgment field were hidden. 

\vspace{1em}

RELATIVE DATE REWRITE RULES

- If the original case uses a relative date/time expression, first resolve it against ref\_date before rewriting.

\end{user_example}
\end{figure*}
\begin{figure*}
\begin{user_example}
- Never output a rewritten booking, reservation, scheduling, planning, attendance, reminder-setting, or time-specific coordination query whose requested target time resolves to the past relative to ref\_date.

- If the validator identified a relative-date error, repair it directly.

- Prefer the smallest natural fix that preserves the case idea:

\quad  - replace an invalid phrase like ``this Wednesday'' with a future-valid phrase like ``next Wednesday''

\quad  - or use an explicit absolute date

\quad  - or shift the request to another natural future-valid time window

- If the context also mentions relative dates, make sure those dates remain coherent with the rewritten query and the timeline.

- Do not keep ambiguous phrasing if it would naturally read as past relative to ref\_date. 

\vspace{1em}

Return JSON only in exactly this format: \{json\_format\}

\vspace{1em}

Field format:

- situation\_type: one of temporal, personal, state

- label: one of conflict, non\_conflict

- query: 1-2 natural sentences, normal assistant request, no hint of hidden reasoning

- context: one concise narrative paragraph containing only the necessary facts

- judgment: 1-2 sentences explaining why the request conflicts with the context or remains compatible with it, using only facts stated in context 

\vspace{1em}

SITUATION TYPE BOUNDARIES

1. temporal

- The main issue concerns timing, scheduling, preparation, travel, sequencing, or coordination.

- For label = ``conflict'', the request clashes with timing-related facts in context.

- For label = ``non\_conflict'', the context includes timing-related facts, but the request still fits them.

- Focus on schedule conflict, compression, timing fit, preparation time, travel feasibility, or coordination burden, not pure impossibility.

- An alter-involved temporal case may use a concrete coordination duty, pickup, caregiving responsibility, or fixed shared plan tied to a specific alter.

2. personal

- The main issue must be grounded in a meaningful personal constraint, boundary, welfare concern, accessibility need, caregiving obligation, or ethically significant commitment tied to ego or a closely related alter.

- Mere dislike, ordinary preference mismatch, or weak taste mismatch is NOT enough.

- A valid personal conflict should make the request meaningfully inappropriate, risky, inconsiderate, or misaligned with a concrete personal constraint.

- Do not treat ``would not enjoy it much'' or ``does not usually prefer it'' as a conflict.

- For label = ``conflict'', carrying out the request would conflict with those personal facts.

- For label = ``non\_conflict'', the request remains compatible with those personal facts or appropriately respects them.

- Do not create a personal case where ego knowingly requests something that directly contradicts ego's own already-known condition, medical limitation, strong established aversion, dietary identity, or firm boundary.

\quad    - Bad example: ego is lactose intolerant and normally avoids dairy at home, but the query is ``Set 

\quad up a weekly milk delivery for my morning coffee.'' This is weak because ego who is lactose 

\quad intolerant would not ask for such request.

\quad    - Bad example: ego generally avoids red meat year-round, but the query is ``Book me a steakhouse 

\quad chef’s tasting tomorrow.'' This is weak for the same reason.

\end{user_example}
\end{figure*}
\begin{figure*}
\begin{user_example}
- Better personal cases usually involve an alter's concrete need or boundary that ego plausibly knows, or a situationally relevant personal constraint that becomes important only after context is added.

- Personal cases must be grounded in concrete persona-linked facts, not generic etiquette, vague discomfort, or broad social impropriety alone.

- If an alter is involved, the case must depend on that alter's specific persona-grounded condition, boundary, need, or responsibility, not merely on the existence of the relationship.

3. state

- The main issue is a world, system, infrastructure, location, or operating condition known by ref\_date.

- For label = ``conflict'', these external conditions make carrying out the request conflicting, risky, wasteful, disruptive, or poorly matched to the situation.

- For label = ``non\_conflict'', the request remains appropriate and workable when those external conditions are taken into account.

- State cases should usually be grounded in conditions already in effect, already announced, already observable, or already scheduled by ref\_date.

- Good examples include temporary access restrictions, route diversions, infrastructure work, building conditions, neighborhood disruptions, public guidance already issued, or announced operating constraints.

- Do not use simple closure, stock failure, outage, or outright access denial as the sole reason.

- Do not use future-only state facts that would become known only after ref\_date.

- An alter-involved state case may use an external condition that becomes relevant because of a specific alter's concrete need, access requirement, or planned involvement.
\end{user_example}
\end{figure*}

\begin{figure*}
\begin{user_example}[frametitle=Prompt for Distractor Generation]
You are a data generation assistant for building distractor context snippets for a conflict-aware retrieval benchmark.

\vspace{1em}

\#\#\# Task Setup

Each benchmark scenario contains multiple cases that all belong to one shared ego-centered world.

Each case contains a situation\_type, a label (conflict or non\_conflict), a user request, a gold context, and a gold judgment.

Your job is NOT to explain the case outcome, justify the label, resolve the case, or create an alternative reason for the judgment. Your job is to generate additional knowledge-base context snippets that could plausibly exist in the same shared world.

These distractor contexts should:

- look superficially relevant to the target query

- resemble natural persona/world knowledge

- remain fully compatible with the entire scenario

- provide little or no useful evidence for determining the correct label of the target case

- provide little or no useful information for reproducing the gold judgment

- provide little or no useful information for producing the final answer

- and preferably be grounded in a concrete harmless anchor point within the timeline window

A distractor context is therefore:

- a plausible knowledge-base fragment in the same world

- related in scene or topic

- but not genuinely useful for deciding whether the target query is conflicting or non-conflicting

- and not materially helpful for explaining why the gold judgment is correct

A strong distractor is retrieval-relevant but decision-irrelevant.
It may look useful enough to retrieve, but it should not substantially reduce the uncertainty needed to judge the case or answer the query. 

\end{user_example}
\end{figure*}
\begin{figure*}
\begin{user_example}

\#\#\# Input Semantic

You will be given: TIMELINE, PERSONA\_EXPANSION, SCENARIO\_CASES, TARGET\_CASE\_INDEX, TARGET\_CASE, N\_DISTRACTORS.

\vspace{1em}

Interpret these fields as follows:

- SCENARIO\_CASES:
  The full set of cases in one shared scenario world.
  You must use all of them to maintain global compatibility.

- TARGET\_CASE:
  The specific case for which you are generating distractor contexts.

- QUERY:
  The user's requested action for the target case.

- SITUATION\_TYPE:
  The case's situation type, such as temporal, personal, or state.
  Use this only to understand the broad scene of the case.
  Do NOT force distractors to mirror this type.

- LABEL:
  The target case is labeled as either conflict or non\_conflict.

- GOLD\_CONTEXT:
  The real situational world state relevant to the target case.

- GOLD\_JUDGMENT:
  The explanation of why the target case is conflict or non\_conflict in the gold world. 

\vspace{1em}

Interpret the target case according to its label:

- If LABEL = conflict:
  distractors must not restate, imply, strengthen, or make it easier to infer the blocking rationale described by the gold judgment.

- If LABEL = non\_conflict:
  distractors must not restate, imply, strengthen, or make it easier to infer the compatibility rationale described by the gold judgment.
  They may be related background details, but they should not materially help confirm that the request is feasible, convenient, available, accessible, or conflict-free. 

\vspace{1em}

\#\#\# Timeline

You are given window\_start, ref\_date, window\_end.

Timeline semantics:

- ref\_date is the current moment when the user makes the request.

- [window\_start, ref\_date] contains facts already observed, learned, confirmed, posted, announced, or in effect by ref\_date.

- (ref\_date, window\_end] may contain future events or conditions only if they are already scheduled, reserved, announced, posted, planned, or otherwise knowable by ref\_date.

A fact or context is valid only if it could already exist in Ego's knowledge base at ref\_date.

Do not invent future facts that would become knowable only after ref\_date. 

\vspace{1em}

\#\#\# Egocentric world model

You are working inside a single coherent egocentric persona world.

- Ego is the central person whose life grounds the scenario and whose knowledge base is being built.

- Alter is a closely related person in Ego's world, such as a family member, close friend, coworker, neighbor, collaborator, or regular contact.

- Every output context must be something that Ego could plausibly know and store by ref\_date.

- If a context involves an alter, include only information that Ego could reasonably know by ref\_date.

- Do not include hidden or private alter information that Ego would not realistically know. 

\vspace{1em}

\#\#\# Global compatibility requirement

Each distractor context must be compatible not only with the target case, but with the entire scenario.

It must not contradict any query, context, judgment, label, or implied world state from any case in the scenario.

It must not create a new conflict for any query in the scenario.

\end{user_example}
\end{figure*}
\begin{figure*}
\begin{user_example}

It must not create a new decisive enabling condition that makes any query obviously answerable either.

Distractor contexts generated for one case must also remain compatible with distractor contexts generated for other cases.

Treat the full scenario as one shared ego-centered world and generate distractors as additional harmless KB fragments for that world. 

\vspace{1em}

\#\#\# Core goal

- The distractor should look superficially relevant to the target query.

- It may share broader scene family, activity family, social sphere, neighborhood, venue type, route family, time-of-day frame, companion type, or lifestyle context.

- But it must be evidentially weak with respect to the target case outcome.

- It must not materially help a model determine whether the correct label is conflict or non\_conflict.

- It must not materially help a model reproduce the gold judgment.

- It must not materially help a model produce the final answer.

- Prefer query-adjacent background context over generic persona biography.

- Distractors may be topically adjacent to the gold evidence, but they must not be decision-useful. 

\vspace{1em}

\#\#\# Temporal anchoring preference

Distractors should not default to timeless generic preferences, gear lists, or broad habits.
Whenever possible, ground each distractor in one concrete harmless anchor point inside the timeline window.

Preferred anchor types include:

- a past observation, note, save, reminder, confirmation, or routine instance in [window\_start, ref\_date]

- a same-day but non-decisive detail on ref\_date

- a future plan, reservation, meetup, ticket, reminder, or venue note in (ref\_date, window\_end] that is already scheduled, announced, reserved, saved, or otherwise knowable by ref\_date

A distractor may be timeless only if a more situationally anchored version would be unnatural.

Whenever possible, make the distractor feel like a concrete world-state note rather than generic biography.

Across multiple distractors, vary the temporal anchoring style. Do not make all distractors timeless.

When natural, aim for a mix such as:

- at least one past-known anchored distractor

- at least one ref\_date-adjacent anchored distractor

- at least one future-known anchored distractor 

\vspace{1em}

\#\#\# Important constraints

- Avoid reusing the specific entities, dates, times, locations, commitments, restrictions, logistical predicates, access conditions, operational facts, or availability details that materially support the gold judgment.

- Distractor contexts do NOT need to be on the same day as the target query.

- They may refer to any fact or plan within the timeline window, as long as it could plausibly be known by ref\_date and remains non-decisive for all scenario queries.

- They may include time-dependent or date-anchored information, provided it is already knowable by ref\_date and does not create, strengthen, or imply any conflict for any scenario query.

- They also must not provide the missing key fact that would make a non\_conflict answer obviously easier.

Each DISTRACTOR must satisfy ALL of the following:

1) Persona-grounded KB style

- Write each distractor as if it were an ordinary knowledge fragment that could have come from the persona expansion or related world-state notes.

- Use details naturally supported by the PERSONA\_EXPANSION.
\end{user_example}
\end{figure*}
\begin{figure*}
\begin{user_example}
- The distractor should feel like background knowledge, not like an explanation or judgment.

2) Scenario-wide compatibility

- The distractor must be fully compatible with the persona expansion, timeline, and all cases in the scenario.

- Do not contradict, negate, or weaken anything in the existing world.

- Do not create friction with another case even if it does not affect the target case.

3) No judgment leakage

- Do NOT restate, paraphrase, imply, strengthen, or weaken the key evidence that determines the correct label or supports the gold judgment.

- Do NOT include context from which a reasonable assistant could much more easily infer the gold judgment.

- For conflict cases:

\quad  - do not restate, paraphrase, imply, or strengthen the blocking rationale
  
\quad  - do not include blocking facts that reproduce the same rejection path
  
- For non\_conflict cases:

\quad  - do not restate, paraphrase, imply, or strengthen the compatibility rationale
  
\quad  - do not provide decisive enabling evidence that directly confirms the request is feasible, available,

\quad accessible, well-timed, or easy to execute
  
\quad  - do not include the missing key fact that would make the correct answer obviously easier

- Do NOT mention or imply:

\quad  - overlap with another commitment
  
\quad  - inability to arrive on time
  
\quad  - unavailability of requested roles or slots
  
\quad  - closures, disruptions, diversions, advisories, or restrictions that directly affect execution
  
\quad  - hard medical, ethical, or physical constraints that directly block a request
  
\quad  - exact feasibility facts whose presence would directly settle a non\_conflict case
  
\quad  - any other fact that makes any scenario query newly impossible, newly trivial to approve, newly trivial to reject, inappropriate, unsafe, or clearly obvious

4) No new outcome shift

- The distractor by itself must not create a separate reason to reject or modify any scenario query.

- It must not add a decisive reason to confidently approve, recommend, or complete a non\_conflict query either.

- If a model saw a scenario query and this distractor alone, the final judgment should remain underdetermined or only weakly affected.

5) Retrieval-candidate quality

- Distractors should resemble plausible KB evidence items, not broad persona biography.

- Strong distractors are often tied to a specific harmless anchor in the timeline window.

- Generate only distractor contexts at this stage. Do NOT output atomic facts.

6) Additional guidance for non\_conflict cases

When the target case is non\_conflict, good distractors are related but non-decisive background details. 

Do NOT generate distractors that directly answer the practical question, such as:

- the exact venue hours needed to confirm availability

- the exact route, timetable, or travel duration needed to confirm feasibility

- the exact reservation policy, access rule, or price detail needed to complete the request

- the exact companion preference or approval that would directly validate the plan

- the exact calendar fact that shows the relevant evening is open

- the exact operational status that would make the answer obvious

7) Diversity

- Distractors should vary in informational focus.

- Avoid near-duplicates and shallow paraphrases.

\end{user_example}
\end{figure*}
\begin{figure*}
\begin{user_example}
- When generating multiple distractors, vary topic focus and temporal anchoring style when natural.

8) Style

- Each distractor should be a short self-contained paragraph.

- Use neutral, factual English.

- Do not include analysis, advice, warnings, or meta-commentary.

- Do not mention that the distractor is non-blocking, weakly evidential, or related to evaluation. 

\vspace{1em}

\#\#\# Generation procedure

For each distractor:

1. Select one harmless anchor point within the timeline window.

2. Make the distractor topically adjacent to the target query.

3. Check the target case label:

\quad   - if conflict, ensure the distractor does not help recover the blocking rationale
\quad   - if non\_conflict, ensure the distractor does not help recover the compatibility rationale

4. Ensure the distractor stays fully compatible with the entire scenario and introduces no outcome-changing evidence.

5. Write a short neutral paragraph describing only the facts. 

\vspace{1em}

\#\#\# Strong vs weak distractor guidance

Weak distractor: ``Regina likes museums and casual dinners.''

Why weak:

- too generic

- not anchored in the timeline window

- reads like broad biography rather than retrievable KB context

Weak in a different way: ``The Wednesday dinner was moved to Thursday, leaving next Wednesday evening free.''

Why weak:

- directly reveals the compatibility rationale

- makes the non\_conflict judgment too easy

Stronger distractor: ``On January 29, Regina saved a note about a small Downtown cafe that stays open later than most neighborhood coffee spots and bookmarked it as a possible post-museum stop for another evening.''

Why stronger:

- grounded in a knowable time point inside the timeline window

- query-adjacent

- harmless and non-decisive

\vspace{1em}

Return ONLY valid JSON in this exact schema: \{json\_schema\}

\#\#\# Field Rule

- context:

\quad  - one concrete narrative paragraph

\quad  - describe only the neutral facts

\quad  - keep it specific, grounded, and persona-linked

\quad  - do not explain the conclusion here

\quad  - all facts in context must be temporally valid under the timeline semantics

\quad  - if the context mentions a future condition, it must already be scheduled, announced, reserved, or 

\quad otherwise knowable by ref\_date

\quad  - whenever natural, ground the paragraph in a specific harmless anchor within the timeline window 

\quad rather than only timeless background preferences

\quad  - the paragraph may look relevant, but it should remain non-decisive for both label inference and 

\quad answer production 
\end{user_example}
\end{figure*}

\begin{figure*}
\begin{user_example}[frametitle=Base System Prompt for Gold and Distractor Context Atomization]
\#\#\# Role

You are an Ego-Centric Knowledge Base Atomizer for conflict-aware RAG systems.
Your task is to convert one source context into 2 to 5 atomic facts that could plausibly be stored in the ego-centric knowledge base by ref\_date.

These facts will later be used by a reasoning model.

Therefore, your output must contain only neutral stored facts, not conclusions, explanations, or logical bridges. 

\vspace{1em}

\#\#\# Multi-hop reasoning objective

The output facts are intended for later evaluation of conflict-aware reasoning.

Each fact should function as an independent neutral evidence unit.

A single fact does not need to reveal the situational relationship by itself.

The relevance of the source context should become identifiable only when multiple facts are integrated with the user query or with other available information.

In other words:

- individual facts should remain plain stored knowledge

- the scenario-level significance should emerge only through later composition and reasoning across facts

Your job is to preserve the evidence needed for that later reasoning step, not to perform the reasoning step now.

\vspace{1em}

\#\#\# Egocentric world model

You are working inside a single coherent egocentric persona world.

- Ego is the central person whose life grounds the scenario and whose knowledge base is being built.

- Alter is a closely related person in Ego's world, such as a family member, close friend, coworker, neighbor, collaborator, or regular contact.

- Every output fact must be something that Ego could plausibly know and store by ref\_date.

- If a fact involves an alter, include only information that Ego could reasonably know by ref\_date.

- Do not include hidden or private alter information that Ego would not realistically know. 

\vspace{1em}

\#\#\# Timeline

You are given: window\_start, ref\_date, window\_end

Timeline semantics:

- ref\_date is the current moment when the user makes the request.

- [window\_start, ref\_date] contains facts already observed, learned, confirmed, posted, announced, or in effect by ref\_date.

- (ref\_date, window\_end] may contain future events or conditions only if they are already scheduled, reserved, announced, posted, planned, or otherwise knowable by ref\_date.

A fact is valid only if it could already exist in Ego's knowledge base at ref\_date.

Do not invent future facts that would become knowable only after ref\_date. 

\vspace{1em}

\#\#\# Source usage

Use the provided source context field as the primary evidence source.

Any auxiliary metadata is provided only to help interpret the context correctly.

Do not output any statement that explains why the request is conflicting, risky, ill-timed, wasteful, inappropriate, harmless, compatible, or suitable.

\end{user_example}
\end{figure*}
\begin{figure*}
\begin{user_example}
\#\#\# Goal

Extract a minimal set of distinct atomic facts from the source context.

The facts should act as KB evidence units that a later reasoning model can combine. 

\vspace{1em}

\#\#\# What an atomic fact is (CRITICAL)

An atomic fact is a single neutral world statement that could be stored independently in the ego-centric knowledge base.

Each fact must:

- contain exactly one claim

- be standalone and understandable by itself

- use explicit names and entities

- be neutral and descriptive

- avoid implications, conclusions, and consequences

- remain useful as a stored KB item even without the scenario explanation

Atomicity granularity rule:

- Interpret ``atomic'' as one indivisible stored knowledge unit, not as the shortest possible text fragment.

- Keep together details that belong to the same real-world assertion when separating them would produce unnatural, weak, or incomplete KB items.

- A fact may include tightly bound attributes of a single event, notice, condition, policy, preference, or item, as long as it still expresses one claim.

- Split facts only when the source contains genuinely distinct claims that could be stored or retrieved independently.

- Do not decompose one coherent source statement into multiple trivial fragments if a single self-contained fact would better represent the stored knowledge.

Redundancy control rule:

- Do not output two facts when one is merely a weaker paraphrase, habitual restatement, or near-duplicate of the other.

- If a stable preference, aversion, policy, or routine can be expressed clearly in one self-contained fact, prefer one fact over multiple overlapping restatements.

Time consistency rule:

- Facts about scheduled events, temporary conditions, posted notices, and date-specific restrictions should include the full relevant calendar date and, when available, the operative time window in the same fact.

- If multiple time details belong to one scheduled event or notice, include them in the same fact rather than splitting them across separate facts. 

\vspace{1em}

\#\#\# Independence and clarity

Each fact must be independently interpretable.

Entity clarity rule:

- Every fact must clearly name the main entity or entities involved.

- Do not omit the subject.

- Do not rely on neighboring facts to identify a person, place, venue, organization, route, event, document, item, or alter.

- Do not use pronouns or demonstratives such as: it, he, she, they, this, that, these, or those.

- Do not use underspecified definite references such as the show, the restaurant, the meeting, the venue, the building, the route, the appointment, or similar forms unless the entity is explicitly identified in the same fact.

- If a specific entity is intended, name it concretely enough for the fact to be understood on its own.

\end{user_example}
\end{figure*}
\begin{figure*}
\begin{user_example}

Reference resolution rule:

- Do not use unresolved temporal, spatial, or situational references.

- Forbidden forms include expressions such as: at that hour, at that time, then, by then, around then, at that point, from there, nearby, locally, in the area, on site, and similar shorthand.

- Even if the source context uses such wording, rewrite it into an explicit, self-contained reference.

- Replace shorthand with the concrete date, time, time window, place, route, venue, or condition it refers to.

- If the reference cannot be resolved cleanly, rewrite the fact to remove the ambiguity rather than copying the original phrasing.

- Do not use comparative or relational time expressions such as earlier, later, before, or after unless the comparison point or reference event is explicitly named in the same fact.

Time anchoring rule:

- Add an explicit date or datetime only when the fact is time-dependent and needs a time anchor for correct interpretation.

- Time-dependent facts include scheduled events, temporary conditions, posted notices, confirmations, reservations, closures, diversions, reminders, saved notes, logged route instances, and other facts whose truth depends on a specific date or time.

- Do not add an explicit date to stable facts that remain clear without one. (e.g., By ref\_date, ego is a scientist who studies microorganisms. -> This fact is true and storable regardless of the date, so it does not need a date anchor.)

- Stable facts include standing routines, long-running traits, durable preferences, persistent conditions, venue properties, and generic travel characteristics.

- If a fact is specifically about a condition or event on ref\_date itself, explicitly write that calendar date.

- Do not use relative expressions such as: today, tomorrow, tonight, this morning, or this evening.

- Do not add a time anchor to a stable fact unless the date is necessary for correct interpretation.

Final self-containment rule:

- Read each fact as if it were retrieved alone from the knowledge base.

- If any entity, time reference, place reference, route reference, or condition would be unclear when read alone, rewrite the fact until it is fully self-contained. 

\vspace{1em}

\#\#\# Lexical variation policy

Do not mechanically copy the wording or expressions of the source context.

Avoid generating nearly identical sentences with merely small wording changes.

Lexical variation and rephrasing are allowed and often preferred.

You may:

- use a near-synonym when the referent remains fully clear

- restate a detail more compactly if no information is lost

You must not:

- replace a clear referent with a vague one

- drop the entity, place, or time anchor needed for independent interpretation

- change the meaning, scope, or strength of the fact

In other words, surface wording may vary, while referential clarity may not vary.
When in doubt, prefer a slightly more explicit fact over a shorter but context-dependent fact. 
\end{user_example}
\end{figure*}
\begin{figure*}
\begin{user_example}
\#\#\# Atomization rule

Extract a minimal set of distinct KB facts from the source context.

- Each fact must contain exactly one claim.

- Treat one coherent event, notice, policy, condition, preference, or schedule as one fact unless the source clearly contains separable claims.

- Do not split one coherent claim into multiple fragments merely to make the output more granular.

- Do not merge claims that could reasonably be stored or retrieved independently.

- If a source sentence contains a primary fact plus a consequence, keep only the primary fact.

- If two candidate facts are near-duplicates or one subsumes the other, keep only the more informative self-contained fact.

- Prefer the fewest facts that preserve all necessary evidence clearly and concretely.

- Do not make any single fact so complete that it already states the scenario-level conclusion on its own. 

\vspace{1em}

\#\#\# Critical bans

Do NOT include:

- any restatement of the query

- any requested time, location, participant, or plan from the query unless that information independently appears as a storable world fact in the source context

- any scenario conclusion

- any explanation of why the fact matters

- any logical bridge connecting multiple facts

- any derived statement that already performs the reasoning step

- any evaluation of feasibility, appropriateness, risk, lateness, inconvenience, harm, compatibility, or suitability

- any statement that the request is possible, impossible, compatible, conflicting, safe, fine, harmless, or problematic

Examples of forbidden outputs:

- The person would need to change an existing plan to carry out the request.

- The person would probably miss the requested start time.

- The requested booking would not fit the known schedule.

- The selected option would not suit the person's physical condition.

- The request conflicts with the known situation.

- The facts together make the request unsuitable.

- The request remains feasible.

- The context does not create any real problem.

Those are reasoning outputs, not KB facts.
\end{user_example}
\end{figure*}

\begin{figure*}
\begin{user_example}[frametitle=Supplementary System Prompt for Gold Context Atomization]
\#\#\# Gold-source specific rule

The source context is a gold case context.

Each gold case includes:

- a situation\_type

- a label, which is either conflict or non\_conflict

- a query

- a context

- a judgment

Use the context field as the primary evidence source.

Use the judgment field only to understand how the context relates to the query.

If label = conflict:

- the judgment explains why the request conflicts with the known situation

If label = non\_conflict:

- the judgment explains why the request remains compatible with the known situation

In both cases:

- Do not copy, paraphrase, compress, or lightly rewrite the judgment field into facts.

- Do not output any statement that explains why the request is conflicting, compatible, risky, safe, ill-timed, harmless, or suitable.

- Output only neutral stored facts from the context.

Completeness requirement for gold:

- The full set of emitted facts should preserve enough neutral evidence that, when combined with the query, a later reasoning model can fully reconstruct the judgment.

- Do not omit a context fact that is necessary for distinguishing why the query is conflict or non\_conflict in this case.

- Preserve all minimally necessary evidence, but keep each individual fact neutral and non-conclusive.
\end{user_example}
\end{figure*}

\begin{figure*}
\begin{user_example}[frametitle=Supplemetary System Prompt for Distractor Context Atomization]
\#\#\# Distractor-source specific rule

The source context is a distractor context.

These facts are not direct gold evidence.

They are related, non-blocking KB facts that may look useful for planning or retrieval.

Preserve such harmless situational anchors when they are part of the distractor context.

Additional distractor constraints:

- Do not transform harmless distractor context into blocking evidence, new obligations, new restrictions, or new execution problems.

- Do not infer any conflict or non-conflict conclusion from the parent case.

- Use only the distractor context itself as evidence for the output facts.
\end{user_example}
\end{figure*}

\begin{figure}
\begin{user_example}[frametitle=Base User Prompt for Gold and Distractor Context Atomization]

Task: Atomize the source context into neutral Knowledge Base facts.

This task is to produce atomic facts for the egocentric knowledge base of \{ego\_name\}.

Timeline: \{timeline\_json\}

Source package: \{source\_json\}

Instructions:

- Extract only the smallest distinct evidence units from the source context.

- Facts must be storable in the ego-centric KB by ref\_date.

- Use the source context as the primary evidence source.

- Facts must be standalone, neutral, and non-inferential.

- Facts must not perform the reasoning step. Facts must not mention why the scenario is good, bad, conflicting, harmless, compatible, or suitable.

- Avoid generating nearly identical sentences with merely small wording changes.

- A later reasoning model should be able to combine the facts with the query to reconstruct the case-level relationship.

- Do not collapse that multi-step reasoning into any single fact.

- Do not include any query restatement. (e.g., The person requests to book a restaurant. -> Do not include this as a fact unless the source context explicitly states it as a known fact.)

- Each fact must contain exactly one claim.

- Every fact must explicitly name its main subject or entity.

- Every time-dependent fact must include an explicit date, datetime, or time window when that anchor is necessary for correct standalone interpretation.

- If multiple time details belong to one scheduled event or notice, include the full relevant time window in the same fact rather than splitting the timing across separate facts.

- Do not omit a time anchor for temporary or date-specific facts.

- Do not force a date onto a stable fact that remains clear without one.

- Do not add a date to a stable fact unless the date is needed for correct interpretation.

- Each fact must remain understandable on its own when separated from the other facts.

- Lexical variation from the source context is allowed and preferred, but the meaning and referents must remain precise and unambiguous.

- Return JSON only.

- No extra keys.

\{extra\_user\_instructions\}
\end{user_example}
\end{figure}

\begin{figure}
\begin{user_example}[frametitle=Supplementary User Prompt for Gold Context Atomization]
Gold-specific instructions:

- Use the context as the primary evidence source.

- Use the judgment field only for interpretation, not for wording.

- Do not copy or paraphrase the judgment sentence into facts.

- Treat conflict and non\_conflict cases in the same way at the atomization level: extract neutral KB facts only.
\end{user_example}
\end{figure}

\begin{figure}
\begin{user_example}[frametitle=Supplementary User Prompt for Distractor Context Atomization]
Distractor-specific instructions:

- Use only the distractor context as evidence for the output facts.

- Facts must not restate the query or the gold judgment.

- Facts must not introduce any new blocking constraint or incompatibility.
\end{user_example}
\end{figure}

\clearpage
\onecolumn
\label{app:method_prompts}

\begin{figure}[t]
\begin{user_method}[frametitle=Prompt for Conflict Planner Agent]
\#\#\# Instruction:

You are a traversal guidance agent for conflict-aware personal assistant retrieval.

You will receive a user request.
Your task is to identify retrieval directions that are likely to uncover useful evidence for downstream conflict reasoning.

\vspace{1em}

* Focus on context directions that may help later retrieval discover: constraints, commitments, routines, preferences, availability, compatibility, dependencies, supporting feasibility evidence, contextual information that may become important when combined with other documents

\vspace{1em}

Useful directions do NOT need to directly imply a conflict.
Indirect or partial contextual evidence may still be important later.

Be selective.
Only include directions that are plausibly useful for retrieving decision-relevant evidence for this specific request.
Avoid generic directions that are not grounded in the request.

\vspace{1em}

* Context directions may involve:

- schedules, timing, recurring routines, or prior commitments

- people involved, relationships, coordination, or availability

- locations, transportation, access, or travel

- preferences, sensitivities, habits, or behavioral tendencies

- physical state, health, mobility, or energy constraints

- resources, reservations, permissions, or required preparation

- contextual information that helps interpret ambiguous or underspecified parts of the request

\vspace{1em}

* Respond with a JSON object: \{json\_schema\}

\vspace{1em}

* Rules:

- Select only 1 to 3 directions. If only one clearly applies, return just one.

- Each direction must be grounded in something specific from the request.

- Do not include speculative directions with no plausible retrieval value.

- Include directions that may support feasibility as well as directions that may reveal constraints.

- key\_anchors must contain exact terms from the request, not generic paraphrases.

- Return valid JSON only. No explanation. No markdown fences.

\vspace{1em}

\#\#\# Request: \{query\}

\#\#\# Reference date: \{ref\_date\}

\vspace{1em}

Identify decision-relevant context directions that would help graph traversal select useful documents for judging this request later. 

Return JSON only.

\vspace{1em}

\#\#\# Response:
\end{user_method}
\end{figure}

\newpage

\begin{figure}[t]
\begin{user_method}[frametitle=Prompt for Multi-View Generator Agent]
\#\#\# Instruction:

You generate retrieval-oriented counter queries for conflict-aware retrieval.

Return JSON with exactly one field: ``counter''.

\vspace{1em}

The reference date is \{ref\_date\}.

When the query contains relative temporal expressions such as today, tomorrow, tonight, this Wednesday, next Friday, or this weekend, interpret them with respect to the reference date when generating counter views.
Do not add extra explanation about the temporal resolution.

\vspace{1em}

Conflict analysis (use this to guide counter query generation): \{conflict\_planner\_result\}

\vspace{1em}

\#\#\# Definitions:

* counter

A list of retrieval queries aimed at finding information that could block, constrain, or conflict with the request.
These should not be simple paraphrases of the original query.

They should help retrieve evidence about conflict-bearing factors such as schedule, availability, timing, commitments, access constraints, coordination constraints, and so on.

If a conflict analysis is provided above, use it to generate counter queries that specifically target those conflict dimensions and anchors. Each counter query should be distinct and targeted.

Avoid near-duplicate counter queries.

\vspace{1em}

* Rules:

- Output valid JSON only

- Do not include explanations

- ``counter'': list of \{num\_counter\} strings

\vspace{1em}

\#\#\# Request: \{query\}

\end{user_method}
\end{figure}

\begin{figure}[t]
\begin{user_method}[frametitle=Prompt for Pre-Hop Filter and Post-Hop Filter Agent]
\#\#\# Instruction:

You are an evidence selection agent for conflict-aware personal assistant retrieval.

You will receive:

1. A user request

2. A list of candidate documents collected through retrieval and graph expansion

Your task is NOT to make the final decision about the request.

Your task is to select the documents that are MOST LIKELY to help a downstream assistant determine whether the request can or cannot be carried out.

\vspace{1em}

* A document may be useful even if:

- it does not explicitly mention a conflict

- it only provides partial information

- it contains habits, routines, preferences, schedules, relationships, locations, or behavioral context

- its relevance may only become clear after combining it with other documents

Focus on retaining documents that may later become important evidence.

\vspace{1em}

* Respond with a JSON object: \{json\_schema\}

\vspace{1em}

* Selection priorities:

1. Documents containing constraints, commitments, obligations, availability, schedules, routines, or resource limitations relevant to the request.

2. Documents containing personal preferences, behavioral tendencies, habits, sensitivities, or compatibility information relevant to the request.

3. Documents that clarify people, places, timing, recurring activities, dependencies, or social relationships mentioned or implied in the request.

4. Documents that may serve as bridge evidence connecting the request to other potentially relevant facts.

\vspace{1em}

* Rules:

- Select exactly \{top\_n\} doc\_ids if at least \{top\_n\} candidate documents are provided.

- Order selected\_doc\_ids from most useful to least useful.

- Prefer documents with concrete, person-specific, time-specific, place-specific, or state-specific information.

- Prefer potentially decision-relevant evidence over topical similarity.

- Do NOT require a document to explicitly prove feasibility or conflict in order to keep it.

- Avoid documents that are purely topical and provide no useful contextual evidence.

- Only use doc\_ids from the provided list.

- Return valid JSON only. No explanation. No markdown fences.

\vspace{1em}

\#\#\# Request: \{query\}

\#\#\# Candidate documents (\{n\_docs\} total): \{docs\_text\}

Select the top \{top\_n\} doc\_ids that are most useful as evidence for deciding whether the request can be carried out.

Respond with JSON only.

\vspace{1em}

\#\#\# Response:
\end{user_method}
\end{figure}

\begin{figure}[t]
\begin{user_method}[frametitle=Prompt for Answer Generator]
\#\#\# Instruction:

You are the personal assistant for \{persona\_name\}.

Today is \{ref\_date\}.

\vspace{1em}

The information below consists of facts you already know about \{persona\_name\} and the world they live in.
Answer the query as their assistant, using those facts when relevant.

\vspace{1em}

Act as an agent who can take actions in that world, coordinate with people and systems there, and speak directly to \{persona\_name\}.
Use what you know to decide whether this request should be carried out.

\vspace{1em}

Write a response of 2 to 4 sentences that clearly states whether the request can or cannot be carried out, and explains why based on the known facts.

\vspace{1em}

Rules:

- Clearly state whether the request can or cannot be carried out.

- Explain why, based only on the provided facts and the query.

- Speak directly as the assistant to \{persona\_name\}.

- Do not mention ``context'', ``provided information'', or ``dataset''.

- Do not refuse with the reason of ``no permission'', assume you are able to take any action as an assistant for \{persona\_name\}.

- Do not use bullet points, JSON, or any labels.

- Keep the response to 2 to 4 sentences.

\vspace{1em}

\#\#\# Context: \{context\_text\}

\#\#\# Query: \{query\}

\vspace{1em}

\#\#\# Response:
\end{user_method}
\end{figure}

\begin{figure}[t]
\begin{user_method}[frametitle=Prompt for Response Evaluation Judge]
\#\#\# Instruction:

You are evaluating a model response for a conflict-aware personal assistant task.

The model response is a free-text answer. It should clearly state whether the user's request can be carried out, and explain why.

You must evaluate two things in order:

\vspace{1em}

Step 1. Decision correctness

- Gold Label indicates the correct decision.

- If Gold Label is ``non\_conflict'', the correct response should proceed, agree, allow, or say the request can be carried out.

- If Gold Label is ``conflict'', the correct response should refuse, warn, block, or say the request should not be carried out as requested.

- If the model response gives the wrong decision, output FAIL.

- If the model response is ambiguous about whether it will proceed or refuse, output FAIL.

\vspace{1em}

Step 2. Rationale correctness

Only evaluate the rationale if the decision is correct.

Compare the model response with the Gold Rationale.

The response does not need exact wording, but it must capture the same core reason.

\vspace{1em}

Use the following labels:

\vspace{1em}

PASS:

The model response conveys the correct decision, and the rationale captures the same core reason as the Gold Rationale.

Minor wording differences are acceptable.

Extra helpful details are acceptable if they do not change or distort the core reason.

\vspace{1em}

WRONG\_RATIONALE:

The model response conveys the correct decision, but the rationale is not fully correct.

This includes cases where the rationale:

- misses key evidence from the Gold Rationale

- is too vague to fully support the decision

- mentions only a secondary reason while missing the main reason

- includes irrelevant or weak justifications

- adds unsupported details that make the reasoning less faithful, while the final decision is still correct

\end{user_method}
\end{figure}
\begin{figure}[t]
\begin{user_method}
FAIL:

The model response conveys the wrong decision.

Also output FAIL if the response is ambiguous about the decision.

Also output FAIL if the decision is correct but the rationale gives the opposite, unsupported, or substantially wrong reason.

\vspace{1em}

Important:

- Decision correctness has priority.

- If the decision is wrong, output FAIL even if some rationale sounds plausible.

- Do not require exact wording.

- Do not penalize conversational phrasing or helpful suggestions.

- Only output one label: PASS, WRONG\_RATIONALE, or FAIL.

- Do not include any explanation.

\vspace{1em}

\#\#\# Gold Label: \{gold\_label\}

\vspace{1em}

\#\#\# Gold Rationale: \{gold\_judgment\}

\vspace{1em}

\#\#\# Model Response: \{answer\_text\}

\vspace{1em}

\#\#\# Label:
\end{user_method}
\end{figure}

\end{document}